\documentclass[journal]{IEEEtran}

\usepackage{hyperref}       
\usepackage{url}            
\usepackage{booktabs}       
\usepackage{amsfonts}       
\usepackage{nicefrac}       
\usepackage{microtype}      
\usepackage{wrapfig}
\usepackage{stfloats}
\usepackage[pdftex]{graphicx}
\usepackage{xcolor}
\usepackage{amsmath} 
\usepackage{multirow}
\usepackage{multicol}
\usepackage{makecell}
\usepackage{colortbl}
\usepackage{float}
\usepackage{enumitem}
\usepackage{cite}
\usepackage[ruled,linesnumbered]{algorithm2e}

\ifCLASSINFOpdf
\else
\fi
\begin{document}
%
\title{Evading Detection Actively: Toward Anti-Forensics against Forgery Localization}

\author{Long~Zhuo, 
	Shenghai Luo,
	Shunquan~Tan*,~\IEEEmembership{Senior Member,~IEEE,}
	Han Chen,
	Bin~Li,~\IEEEmembership{Senior Member,~IEEE,}
	and~Jiwu~Huang,~\IEEEmembership{Fellow,~IEEE}
	\thanks{*Corresponding author: Shunquan Tan.}
	\thanks{All of members are with the Guangdong Key Laboratory of Intelligent Information Processing, Shenzhen Key Laboratory of Media Security, Shenzhen Institute of Artificial Intelligence and Robotics for Society, China (email: zhuolong, luoshenghai, chenhan@email.szu.edu.cn; tansq, libin, jwhuang@szu.edu.cn). }
	
	\thanks{S. Luo and S. Tan are with College of Computer Science and
		Software Engineering, Shenzhen University, Shenzhen 518060,
		China } 
	\thanks{This work was supported in part by NSFC~(U19B2022, 62272314, U22B2047), Guangdong Basic and Applied Basic Research Foundation~(2019B151502001), and Shenzhen R\&D Program~(JCYJ20200109105008228).}
}
\maketitle

\begin{abstract}
	Anti-forensics seeks to eliminate or conceal traces of tampering artifacts. Typically, anti-forensic methods are designed to deceive binary detectors and persuade them to misjudge the authenticity of an image. However, to the best of our knowledge, no attempts have been made to deceive forgery detectors at the pixel level and mis-locate forged regions. Traditional adversarial attack methods cannot be directly used against forgery localization due to the following defects: 1) they tend to just naively induce the target forensic models to flip their pixel-level pristine or forged decisions; 2) their anti-forensics performance tends to be severely degraded when faced with the unseen forensic models; 3) they lose validity once the target forensic models are retrained with the anti-forensics images generated by them. To tackle the three defects, we propose SEAR (Self-supErvised Anti-foRensics), a novel self-supervised and adversarial training algorithm that effectively trains deep-learning anti-forensic models against forgery localization. SEAR sets a pretext task to reconstruct perturbation for self-supervised learning. In adversarial training, SEAR employs a forgery localization model as a supervisor to explore tampering features and constructs a deep-learning concealer to erase corresponding traces. We have conducted large-scale experiments across diverse datasets. The experimental results demonstrate that, through the combination of self-supervised learning and adversarial learning, SEAR successfully deceives the state-of-the-art forgery localization methods, as well as tackle the three defects regarding traditional adversarial attack methods mentioned above.

\end{abstract}


%
\IEEEpeerreviewmaketitle

\section{Introduction}

In recent years, the ease of image tampering with the availability of editing software has led to serious societal problems.~\cite{verdoliva_jstsp_2020}. Image forensics has emerged as a solution to reveal such manipulations. It has two levels: image-level and pixel-level. Image-level forensics detects tampering in an entire image, while pixel-level forensics, also known as forgery localization, goes further to identify tampered regions within an image. Unlike image-level forensics that provides a binary true/false label, pixel-level forensics generates masks that indicate the authenticity of each pixel in the target image. In particular, pixel groups masked with zero indicate non-tampered regions, while those masked with one represent the tampered regions.

The most common types of content-tampering are splicing, copy-move and removal. Traditional feature-based image forgery localization methods focused on a specific type of tampering~\cite{lyu_ijcv_2014,cozzolino_tifs_2015,fan_icip_2015,han_jei_2016,li_neo_2017} with feature-based methods. In recent years, general deep-learning based forgery localization methods~\cite{zhou_cvpr_2018,bappy_tip_2019,wu_cvpr_2019,hu_eccv_2020,zhuo_tifs_2022} have demonstrated superior performance over those traditional feature-based ones.
Among them, the first notable one is RGB-N~\cite{zhou_cvpr_2018}
built upon a two-stream Faster R-CNN network. Another approach is a separate encoder-decoder based
forgery localization framework proposed by Bappy et
al.~\cite{bappy_tip_2019}. Recently,  Mantra-net, proposed by Wu et al.~\cite{wu_cvpr_2019}, creates a unified feature representation for directly localizing forgery regions without extra preprocessing and postprocessing. SPAN, proposed by Hu et
al.~\cite{hu_eccv_2020},  utilizes a pyramid structure of
local self-attention blocks to model the relationships between patches
on multiple scales of the target image. Most recently, SATFL, proposed by Zhuo et al.~\cite{zhuo_tifs_2022}, presents a forgery attention mechanism and self-adversarial training to achieve more robust performance. Regarding the united image forgery detection and localization framework, Guo et al.~\cite{guo_cvpr_2023} present a hierarchical fine-grained framework and Guillaro et al.~\cite{guillaro_cvpr_2023} introduce noise-sensitive fingerprint to neural network.

As the adversary of image forensics, image anti-forensics is the art of hiding tampering traces to deceive image forensics methods. It modifies the manipulated images to fool the forensics algorithms. In the multimedia security community, image anti-forensics and forensics have mutually promoted each other as opponents. As far as we know, all existing image anti-forensic algorithms can only be used
to deceive a target detector at image level. Early proposals try to obscure any evidence of post-processing operations which can be used as hints of
possible tampering manipulations, such as resampling, double JPEG
compression, median filtering, and contrast
enhancement~\cite{kirchner_tifs_2008,kwok_iwdw_2011,stamm_tifs_2011,wu_icassp_2013,lu_ijdcf_2013,ravi_spl_2015,kim_spl_2017,luo_eusipco_2018,shen_iwdw_2019,singh_mssp_2019}. Adversarial training policy has enable proposals in
\cite{wu_apasipa_2019,wu_cs_2021,xie_tcsvt_2021} to implement multi-operation image anti-forensics. Similarly, proposals in
\cite{ding_tmm_2021,peng_tcsvt_2022} attempt to evade the detection of facial image forgery using adversarial training.

From the perspective of anti-forensics, pixel-level anti-forensics is much more challenging than its image-level counterpart. We define pixel-level anti-forensics as eliminating the tampering traces and cheating forgery localization model. With advanced data augmentation techniques, state-of-the-art forgery
localization methods such as SPAN~\cite{hu_eccv_2020} and
SATFL~\cite{zhuo_tifs_2022} still maintain their performance when
confronted with those forged images after post-processing.

However, as previously stated, all the state-of-the-art forgery
localization methods are deep learning based, and they are
inevitably vulnerable to adversarial attack, first proposed by
Goodfellow et al.~\cite{goodfellow_arxiv_2014}. Adversarial attacks deceive deep-learning models by adding small, imperceptible changes to target images, resulting in machine recognition errors that are undetectable to the human eye. These attacks can be classified into three main categories: gradient-based~\cite{goodfellow_arxiv_2014,moosavi_cvpr_2016,papernot_esp_2016,kurakin_arxiv_2016},
optimization-based
attacks~\cite{carlini_arxiv_2017,chen_ais_2017,cheng_arxiv_2018,dong_cvpr_2018}
and GAN-based attacks~\cite{xiao_arxiv_2018,bai_icip_2021}. 

Considering the comprehensive experiments, we apply three effective but non-learnable attack methods and an advanced and learnable GAN-based attack method for experimentation. In particular, we experiment with four representative adversarial attack algorithms: Fast Gradient Sign Method (FGSM) proposed by Goodfellow et al.~\cite{goodfellow_arxiv_2014}, which creates adversarial samples based on the gradients of the targeted network; Basic Iterative Method (BIM) proposed by Kurakin et al.~\cite{kurakin_arxiv_2016}, which improves FGSM in an iterative manner; Momentum Iterative Method (MIM) proposed by Dong et al.~\cite{dong_cvpr_2018}, which utilizes momentum to enhance adversarial attack; and adversarial examples generative adversarial networks (AdvGAN) presented by Xiao et al.~\cite{xiao_arxiv_2018}, which constructed a learnable paradigm with a generator deep network and a classified network to yield the adversarial examples. FGSM is a pioneering adversarial attack that generates perturbation maps by maximizing the detection models' loss. BIM improves FGSM by adopting a new loss function. MIM performs an iterative attack using gradients and a decay factor. The three powerful but non-learnable attack algorithms are extensively applied to deceive classification models, yielding imperceptible perturbations that achieve high attack success rates. Unlike the non-learnable methods, AdvGAN take the advantages of GANs and applies a CNN-based generator to cheat the classifier by predicting the adversarial example maps. The learnable method does not need the target model while generation and is more flexible than non-learnable methods with higher attack successful rate. We provide a comprehensive study of these practical attack methods for anti-forensics against forgery localization.

\begin{figure}
	\centering
	
	\includegraphics[width=1\linewidth]{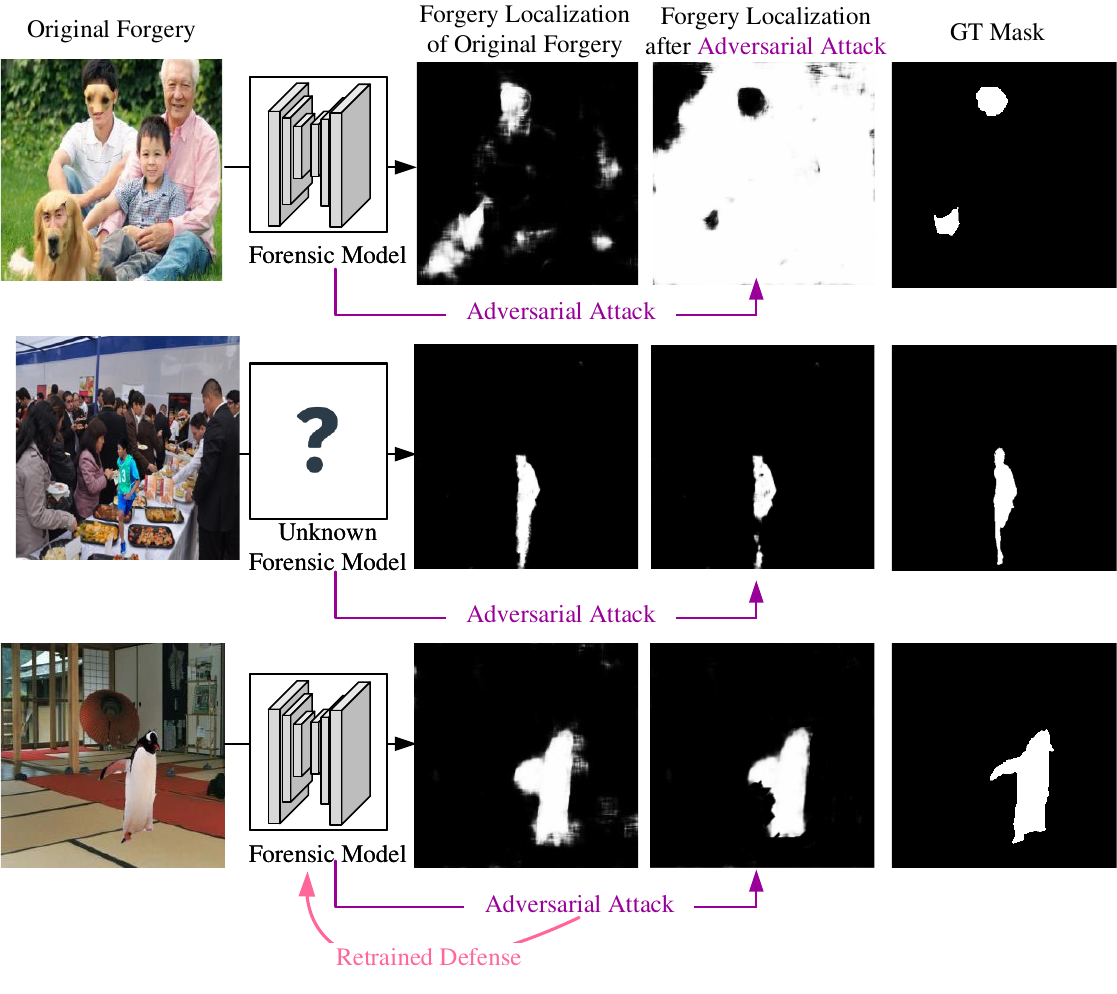}
	\caption{Illustration of the three challenges of the adversarial attack for image anti-forensics at pixel-level. The first row shows the zero-one reversal phenomenon. The second row shows the generalization issue (facing unseen models). The third row shows the failure issue against the retrained defense.
    The first column is the original forged images, the second column is the prediction by the forensic models, the third and the fourth column are the forgery localization models' predictions of original forgery and the attacked images by the conventional adversarial attacks~\cite{goodfellow_arxiv_2014,kurakin_arxiv_2016,dong_cvpr_2018,xiao_arxiv_2018}. The last column is the ground-truth masks of the tampered regions. 
	}
	\label{reversed}
\end{figure}

Nonetheless, as illustrated in Fig~\ref{reversed}, the conventional adversarial attacks cannot be utilized to effectively combat forgery localization methods.

Firstly, adversarial attacks are prone to naively conduct zero-one
reversal to the predicted mask. Serious-minded forensics experts can
still figure out suspicious tampered areas from the reversed
predicted mask.

Second, adversarial attacks are most effective when the attacker has full access to the target model's structure and parameters, a situation known as the white-box setting. However, their effectiveness decreases sharply when the attacker is confronted with unknown forgery localization models.

And thirdly, adversarial attacks tend to fail when confronted
with a sensible target. A sensible forgery localization model can
usually successfully defend itself using attacked tamped images as
fine-tuned samples.

To address the above three issues, we propose a novel
anti-forensic system, SEAR (Self-supErvised Anti-foRensics),
to evade detection of state-of-the-art deep-learning based forgery
localization models. Specifically, with SEAR, the following two main targeted solutions are proposed accordingly:

Firstly, we introduce self-supervised learning to confine adversarial perturbations to tampered areas. We set a pretext task with ground-truth masks to make the network learn the knowledge of the location information of tampered regions. This allows our model, a concealer, to adaptively generate perturbations that modify the tampered images locally, avoiding zero-one reversal.

Furthermore, we incorporate a dilated convolutional group into our proposed concealer network to smooth the tampering artifacts, which are often manifest in the high-frequency domain~\cite{zhou_cvpr_2018,wu_cvpr_2019,zhuo_tifs_2022}. This enables the concealer to effectively hide high-frequency traces.

To address the second and third drawbacks, we construct an adversarial learning scheme with a concealer and a supervisor. The supervisor is a well-trained forgery localization model with advanced network structures, while the concealer is taught to hide the traces of the forged images from the supervisor. Our intuition is that a great supervisor can mentor an excellent concealer, just as with human instruction. This allows our concealer to fight against other state-of-the-art forgery localization networks. To reach convergence, we use adversarial training to train our concealer in an end-to-end manner. Unlike AdvGAN~\cite{xiao_arxiv_2018}, the supervisor is a pixel-level detector instead of an image-level detector. Through competition with the supervisor's dynamical parameters, our trained concealer can deceive unseen or pre-defense forensic models.

Please note that the ground-truth masks and the supervisor are only used in the training phase; our concealer only requires tampered images to conduct evasion.


To summarize, our contributions can be concluded as follows:
\begin{itemize}
	\item We formally define a pixel-level anti-forensics task and propose SEAR, a jointly pixel-level anti-forensic model, against forgery localization for multiple tampering techniques.
	\item 
	The proposed SEAR is trained in an end-to-end manner, utilizing a novel self-supervised learning with a pretext task and adversarial training scheme with a well-trained supervisor.
	\item Extensive and detailed experiments have demonstrated that our SEAR overcomes the mentioned three challenges and deceives the advanced forgery localization methods successfully.
\end{itemize}

\begin{figure*}[t]
	\centering
	
	\includegraphics[width=0.9\linewidth]{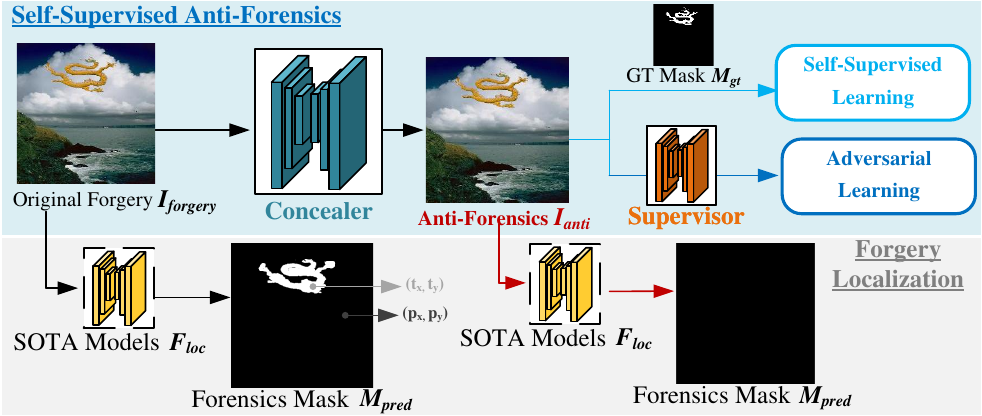}
	\caption{The key concept of pixel-level anti-forensics task and our proposed anti-forensics against forgery localization. The top figure illustrates self-supervised and adversarial learning, consisting of a concealer and a well-trained supervisor. The original forged image with its ground-truth mask (GT mask) are used to train the concealer, while the concealer is trained to compete with the supervisor. During inference, only the original forged image is fed into the concealer, generating an anti-forensic image. The bottom figures illustrate the difference in the predicted results between the original forged image and the anti-forensic image. The left one shows that the state-of-the-art (SOTA) models for forgery localization predict the accurate mask of the original forgery. In contrast, the right one shows that SOTA models predict the blank mask without the location information of tampered regions in the anti-forensic image.  }
	\label{overall}
\end{figure*}

\section{Preliminaries}
\subsection{Forgery Localization}
Forgery localization only focuses on content-aware manipulations, such as splicing, removal and copy-move. Given an image tampered with the content-aware manipulations, the forgery localization methods~\cite{zhou_cvpr_2018,bappy_tip_2019,wu_cvpr_2019,hu_eccv_2020,zhuo_tifs_2022} detect the locations of the tampered regions. These methods predict a mask with the same dimensions (height and width) $(H, W)$ as the corresponding tampered image. Commonly, each pixel of the output mask is binary and classified into two classes, valued at 0 or 1. The coordinate $(p_x, p_y)\in (H, W)$ with value 0 represents that the coordinate $(p_x,p_y))$ is a pristine pixel, while with value 1 refers to a tampered pixel $(t_x,t_y)$. For better visualization, the collections of tampered pixels $\{(t_x,t_y)\}$ in the mask are drawn in white while the pristine collections $\{(p_x,p_y)\}$ are drawn in black. Here, we describe the forgery localization model as $F_{loc}$, as shown in Fig.~\ref{overall}.

To evaluate the performance of the forgery localization methods, several metrics are used, such as AUC and $F_1$ score. $F_1$ score exhibits greater sensitivity compared to the AUC metric when the forgery localization method is under attack~\cite{zhuo_tifs_2022}. Thus, we adopt $F_1$ score as the basic metric in this paper to evaluate the performance of the forensic methods. The calculation of $F_1$ score can be defined as,
\begin{equation}
F_1 = \frac{2\times (Precision \times Recall)}{(Precision+Recall)},
\end{equation}
where $Precision$ denotes correct positive predictions relative to total positive predictions, and $Recall$ denotes correct positive predictions relative to total actual positives. Note that $F_1$ score is calculated for the accuracy of every pixel in the task of forgery localization, and it is also called pixel-level $F_1$ score.

\subsection{Self-Supervised Learning}

Self-supervised learning is a type of unsupervised learning that trains models from data without labels, instead relying on related knowledge. This approach was developed to find a paradigm for learning visual features from unlabeled data~\cite{jing_pami_2020}, which are generated by the machine based on certain attributes. The self-supervised learning method prescribes a pretext task for the neural network to solve, and uses pseudo labels for this task. The neural network can then learn visual features while being trained on the pretext task. The knowledge of these visual features, such as locations, corners, and textures, can then be transferred to downstream tasks.

Our goal for pixel-level anti-forensics is to create anti-forensic images that can effectively conceal tampering traces and deceive detection methods. To achieve this without compromising image quality, we need a concealer that can modify tampered regions while leaving pristine regions unaltered. However, since there are no existing anti-forensic images available for supervised learning, we must employ the existing knowledge. To do this, we introduce a pretext task and generate pseudo labels. The self-supervised learning process then trains the concealer to limit perturbations to the tampered regions.

\subsection{Adversarial Learning}

Adversarial learning proposed by Goodfellow \textit{et al.}~\cite{goodfellow_nips_2014} is a zero-sum game in which two neural networks compete during training. The two networks are opposites striving to achieve the targets. Typically, a generator network produces synthesized data to deceive a discriminator network that is designed to detect the authenticity of the data. Initially, the discriminator network only predicted the real or fake label at the image level. Adversarial learning has achieved remarkable success in image-level anti-forensics~\cite{wu_apasipa_2019,wu_cs_2021,xie_tcsvt_2021}.

\section{Our Proposed SEAR}

In this section, we first describe the anti-forensics problem against forgery localization, and then we provide a detailed description of the proposed SEAR, a joint learning system.

\subsection{Problem Statement}
Pixel-level anti-forensics task aims to conceal the tampering traces by modifying the regional information without destroying the visual effects. The modifications of the local regions are implemented by adding perturbation, the change of each pixel, into the original forged images. In particular, the anti-forensic image can be formulated as

\begin{align}
	I_{anti} &= I_{forgery} + \mathcal{P}_{perturb}(I_{forgery})\notag \\
	&= I_{forgery} + I_\triangle,  
\end{align}
 
where the $I_{forgery}$ denotes the original forged image, and $\mathcal{P}_{perturb}$ is the function to generate the perturbation map $I_\triangle$. This task is to explore a method $\mathcal{P}_{perturb}$ to add the perturbation into the original forged image. For convenience, we define the forgery localization models as $F_{loc}$ and their results, the predicted forensic masks, as $M_{pred}$. Intuitively, two goals are intuitively necessary for $I_{anti}$ to achieve. Firstly, it is important to maintain visual similarity with $I_{forgery}$, meaning that $\mathcal{P}_{perturb}$ needs to avoid significant perturbation. Secondly, it is required to deceive $F_{loc}$. To achieve this, SEAR uses two losses during back-propagation: an restraint loss $Loss_{self}$ and an adversarial loss $Loss_{adv}$.

Fig.~\ref{overall} illustrates the difference between the predicted mask of the original forgery $F_{loc}(I_{forgery})$ and the predicted mask of the anti-forensics $F_{loc}(I_{anti})$. Notably, $I_{anti}$ is visually indistinguishable from $I_{forgery}$. However, the forensic mask $M_{pred}$ of $F_{loc}(I_{forgery})$ predicts the accuracy locations of the tampered regions while $M_{pred}$ of $F_{loc}(I_{anti})$ predicts that there are no tampered regions. This demonstrates that the forensic models have failed to trace the forged information. Next, we introduce SEAR for details to adaptively restrain the redundant perturbation yet fool the forensic models.

\subsection{Overview of SEAR}

\begin{figure*}
	\centering
	\includegraphics[width=0.9\linewidth]{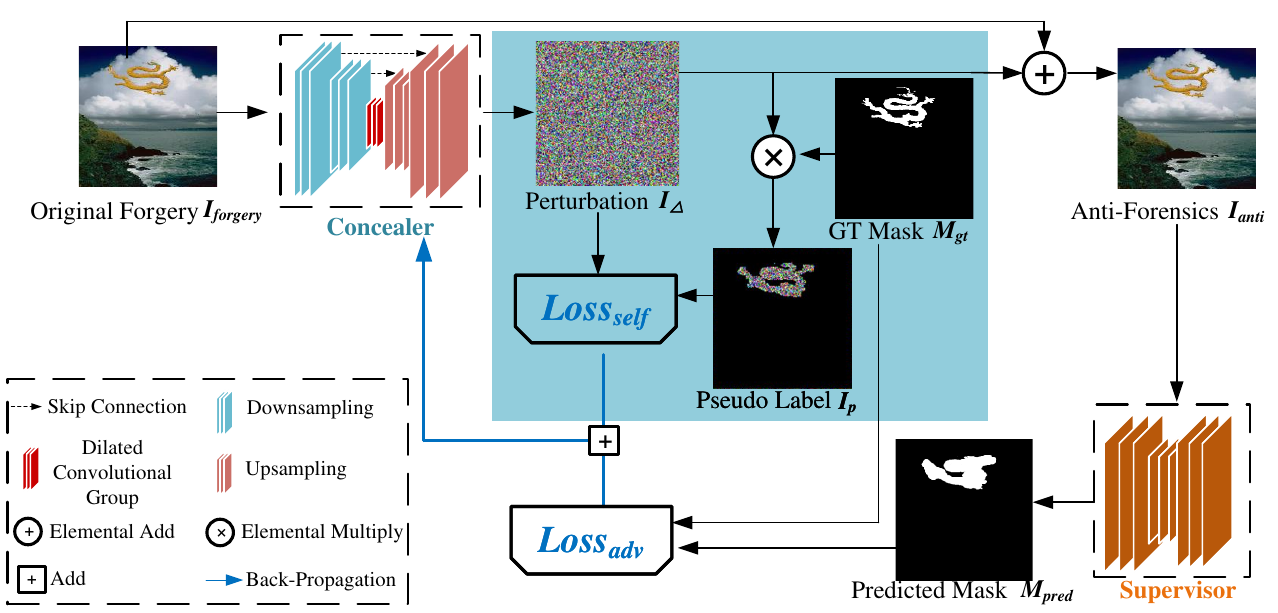}
	\caption{Illustration of the architecture of SEAR. The concealer is constructed by encoder-decoder structures in U-Net style. The original forged image is first fed into the concealer network to generate a perturbation map. The perturbation map is multiplied with the ground-truth mask and then yield the pseudo labels for self-supervised learning. The perturbation map produces the self-adversarial training loss $L_{self}$. Then, the anti-forensics is fed into the well-trained supervised, and we calculate the adversarial loss $L_{adv}$. Two losses are fused to the back-propagation for the system.}
	\label{net}
\end{figure*}

In this work, we address the anti-forensics against forgery localization by introducing a joint learning system of self-supervised and adversarial learning, called Self-supErvised Anti-foRensics (SEAR). SEAR optimizes deep-learning based neural networks using back-propagation~\cite{rumelhart_nature_1986} to adaptively restrain the perturbation to the tampered regions while maintaining minimal visual quality loss. The key idea of our pipeline is presented in Fig.~\ref{overall}, and further details about SEAR are provided in Fig.~\ref{net}. SEAR consists of two deep neural networks: a concealer network and a supervised network. Based on the two networks, it utilizes a combination of self-supervised learning and adversarial learning to extract feature knowledge of tampered locations for the concealer and make it capable of coping with complex and changing forensic methods in practice.

In self-supervised learning, we set a pretext task designed to encourage the model to learn how to restrict perturbations. The ground-truth masks are used to generate pseudo labels for this task. Meanwhile, a dilated convolutional group is applied to the concealer to smooth out any high-frequency traces. In adversarial learning, the supervisor network is continually trained against the concealer. By combining these two methods, SEAR enables the concealer to be back-propagated and learn to effectively hide tampering artifacts.

As illustrated in Fig.~\ref{net}, SEAR is trained to fuse the perturbation map $I_{\triangle}$ adaptively into $I_{forgery}$ and remove tampered traces. To achieve this, we propose a self-supervised learning scheme (outlined in Sect.~\ref{sec:self}), along with adversarial training (described in Sect.~\ref{sec:adv}). The learning system comprises two deep neural networks: the concealer network, which works to eliminate the traces detected by the deep forensic model in the tampering regions, and the supervisor network, which imitates the flexible detection environment for concealer training. 

\subsection{Self-Supervised Learning}
\label{sec:self}
Self-supervised learning offers a powerful solution to the anti-forensics task, as it enables us to train deep networks without the need for manual or mechanical annotation. In this anti-forensics task, there is no labeled data or targeted images to guide the deep network under supervised learning since there are only forged images and their ground-truth masks without anti-forensic images. Consequently, we have endeavored to explore a novel learning paradigm that will strategically eliminate tampering artifacts automatically.

In particular, we set up a pretext task for the concealer to learn to restrict perturbations to the manipulated regions. Intuitively, tampering traces exist in the tampered regions, so the perturbation should smooth these regions while avoiding too much impact and minimizing any potential effects on the pristine areas.

The ground-truth masks contain information regarding the locations of tampered regions, which serve as a crucial reference for our approach. We leverage this knowledge to guide the concealer in constraining any perturbations within the tampered regions. 
We illustrate the pretext task in the blue box of Fig.~\ref{net}. As shown, the concealer predicts the perturbation map $I_\triangle$ of a original forged image $I_{forgery}$. To guide the concealer constrain the perturbation into the tampered regions, we first yield the pseudo label map $I_p$ map by elementally multiplying the ground-truth mask $M_{gt}$ of $I_{forgery}$ and $I_\triangle$. $I_p$ is the target of the pretext task, indicating that we aim to modify the tampering traces while disregarding the pristine regions.

The objective loss function of the pretext task can be written as:

\begin{equation}
	\label{loss_pretext}
Loss_{pretext} = \sum_{(i,j)}||I_{p}^{(i,j)}-I_{\triangle}^{(i,j)}||.	
\end{equation}

During training, the original forged images are fed into the concealer, which generates the corresponding perturbation maps $I_\triangle$. To ensure the visual effect of the final results, it is expected that the values of $I_\triangle$ remain low. Therefore, we employ a hinge loss~\cite{rosasco_nc_2004} to restrict the values of $I_\triangle$. This loss can be calculated as follows:
\begin{equation}
	\label{loss_self}
	Loss_{hinge}= \sqrt{\sum_{(i,j)} (I_{\triangle}^{(i,j)})^2},
\end{equation}
where $(i,j)$ is the coordinate of each pixel of $I_{\triangle}$.

The overall loss function of self-supervised learning is the combination of two objective loss functions and can be expressed as follows:
\begin{equation}
	\label{loss_self}
	Loss_{self} = Loss_{pretext} +Loss_{hinge}.
\end{equation}

This training process does not need the labeled results as the targets for supervision of $I_{anti}$, instead it uses the ground-truth masks for yielding pseudo labels to extract visual features. Thus, this constitutes a self-supervised learning paradigm. By following this strategy for self-supervised learning, the tampering traces' explicit locations are made available, yielding two benefits to the concealer. Firstly, the concealer learns how to restrain the perturbation to the tampered regions. Secondly, the features of the tampered traces are extracted to aid in the smoothing of high-frequency artifacts.

Please note that the ground-truth masks are only utilized during training in pretext task. After training, the concealer network operates without fine-tuning and generates anti-forensic images without any auxiliary mask. For deceiving the detector, adversarial training is proposed to train the concealer network to facilitate deception.

\subsection{Adversarial Training}
\label{sec:adv}

Utilizing self-supervised learning, our concealer can extract tampering trace features and generate a perturbation map to effectively conceal them. Considering the rapid development of the flexible forensic models, we employ a supervisor to guide the concealer for fooling the forensic models. 

To achieve this, a concealer network and a supervisor network are combined for adversarial training. The adversarial training has been used for fooling the classification models~\cite{xiao_arxiv_2018,xie_tcsvt_2021,peng_tcsvt_2022}. Moreover, we make a further improvement to deceive the forensic models at pixel level.

The supervisor is a well-trained forgery localization model, SATFL~\cite{zhuo_tifs_2022}, which predicts the accuracy masks of the corresponding original forged images. Specifically, the anti-forensics images $I_{anti}$ generated by the concealer are used to pass through the supervisor network to generate the predicted masks. The predicted masks $M_{pred}$ are used for calculating the adversarial loss. We adopt binary cross-entropy loss function as our adversarial loss, which can be formulated as,
\begin{equation}
	\label{loss_adv}
	Loss_{adv} = -BCE(M_{pred},M_{gt}),
\end{equation}
where $BCE$ is the binary cross-entropy loss function calculating the distance between the predictions and the labels at pixel-level, $M_{gt}$ denotes the ground-truth masks of the original forged images, and $M_{pred}$ is the supervisor's predictions of the anti-forensic images. The purpose of this loss function is to maximize the distance of $M_{pred}$ and $M_{gt}$ during the back propagation to fool the powerful supervisor.

The adversarial training involves a competitive dynamic between the concealer and the supervisor. Following the update of the concealer's gradients, the supervisor is trained with the loss $-Loss_{adv}$. It indicates that the supervisor continues to train under the attack of the concealer to minimize the distance between $M_{pred}$ and $M_{gt}$. With continuous adversarial training, the supervisor learns to defend against the concealer's attacks and develop a more potent forensic model. The robustness of the supervisor empowers it to educate the concealer on how to combat intricate and versatile forensic models in real-world.

\subsection{Model Architecture}

In the preceding discussion, we utilize two deep networks for constructing our system, namely the concealer network and the supervisor network. In the following, we provide a detailed description of the two networks.

\subsubsection{Concealer Network}
\label{sec:concealer}

As shown in Fig.~\ref{net}, the concealer network is designed to conceal the artifacts in the tampered regions of a forgery image and produce an anti-forensics image with minimal negative visual impact.

To achieve this goal, we adopt the common encoding-decoding architecture and construct the concealer network using VGG-style blocks, specifically the VGGBlock~\cite{simonyan_arxiv_2014}. We choose this type of block due to its stable convolutional behavior, which helps to maintain the overall image quality. Through our ablative experiments, shown in Sect.~\ref{sec:abl-vgg}, we validated the effectiveness of a single VGGBlock in our proposed system.

Each VGGBlock comprises three stacked convolutional layers of $3\times3$ kernel, and can be expressed as follows:
\begin{equation}
	VGGBlock(x) = Conv_{3\times 3}(Conv_{3\times 3}(Conv_{3\times 3} (x))),
\end{equation}
where $x$ is the input, $Conv_{3\times 3}$ denotes convolutional operation with the kernel size $3\times3$. A max pooling operation is also used for downsampling the feature maps' size. A VGGBlock and a max pooling layer are combined into a downsampling block. Mathematically, this can be expressed as:
\begin{equation}
	Downsampling(x) = MaxPooling(VGGBlock(x)),
\end{equation}
where $MaxPooling$ is a max pooling operation to reduce the resolutions of feature map.

To restore the size of the feature maps, the upsampling block is employed. A single upsampling block can be formulated as
\begin{equation}
	Upsampling(x) = Conv_{3\times 3} (Upscaling(x)),
\end{equation}
where $Upscaling$ magnifies the size of the feature map $x$ by a factor of two horizontally and vertically. The encoding process consists of the cascade downsampling blocks and the decoding process consists of the cascade upsampling blocks.

Regarding the previously mentioned dilated convolutional group, it is utilized to smooth high-frequency traces with large receptive fields. Dilated convolutional layers are a type of convolutional filters that perform group convolutions in parallel across the input channels, and they have been shown to be effective in capturing contextual information~\cite{zhuo_tifs_2022}. Specifically, we employ a series of dilated convolution layers with dilation rates of 2, 4, 8, and 16 to link the encoding process and the decoding process.

Furthermore, we adopt the UNET architecture~\cite{ronneberger_miccai_2015}, which has been widely used in image processing tasks and is known for its effectiveness in preserving image details while maintaining a high-level semantic understanding, for enhancing the network representations. In particular, skip connections are used to connect the corresponding outputs of the encoding and decoding module. In this way, the concealer can generate the perturbation map to smooth the tampering traces.

\subsubsection{Supervisor Network}

The supervisor takes a forgery image as input and predicts a binary mask. We employ SATFL~\cite{zhuo_tifs_2022}, a powerful forgery localization model, as our supervisor. Our supervisor network comprises four key components: a Channel-Wise High Pass Filter block, a forgery attention module, VGG-style blocks, and a series of dilated convolution layers. To expedite the training process, we use only the refined net of SATFL, rather than the coarse-to-fine network. Due to space limitations, we do not detail SATFL here. Please refer to~\cite{zhuo_tifs_2022} for specific definitions of SATFL.

\begin{algorithm}[h]
	
	\caption{Joint Learning System of SEAR}
	\label{alg}
	\KwIn{ The concealer network $\mathcal{P}_{perturb}$; \\
	The supervisor network $F_{loc}$. \\

	}
	\KwOut{Anti-forensic images $I_{anti}$.}

	\BlankLine
	Well-train $F_{loc}$; \\
	Initial the parameters of $\mathcal{P}_{perturb}$;\\

	\While{not converged}{
		Freeze the parameters of $F_{loc}$; \\	
		Sample batch data pairs ($I_{forgery}$, $M_{gt}$);\\
		Generate perturbation maps $I_\triangle = \mathcal{F}_{perturb}(I_{forgery})$; \\
		Yield pseudo labels $I_p = I_\triangle \bigotimes M_{gt}$\\
		Synthesize anti-forensic images $I_{anti} = I_{forgery}+I_\triangle$; \\
		Compute attack loss $Loss_{self}$ using Eq.~\ref{loss_self}; \\
		Predict the masks of the anti-forensic images $M_{pred} = F_{loc}(I_{anti})$; \\
		Compute adversarial loss $Loss_{adv}$ using Eq.~\ref{loss_adv}; \\
		Combine the losses to $Loss_{Concealer}$ using Eq.~\ref{loss_c}; \\ 
		Update the parameters of $\mathcal{P}_{perturb}$ by minimizing $Loss_{Concealer}$; \\

		\BlankLine
		Unfreeze the parameters of $F_{loc}$; \\	
		Combine loss $Loss_{Supervisor}$ using Eq.~\ref{loss_s}; \\ 
		Update the parameters of $F_{loc}$ by minimizing $Loss_{Supervisor}$;

	}
\end{algorithm}

\subsection{Joint Learning System}

We combine two deep networks and two learning strategies into a joint learning system, as depicted in Alg.~\ref{alg}. We train the supervisor network $F_{loc}$ in the forgery localization setting, and freeze its parameters in each training iteration to prevent any negative impacts on the concealer. The concealer, denoted by $\mathcal{P}{perturb}$, generates perturbation maps $I_\triangle$ for the input forged images. The anti-forensic images can be calculated as the pixel-wise sum of $I_\triangle$ and $I_{forgery}$. The perturbation maps are then multiplied element-wise with ground-truth masks $M_{gt}$, and the results are added to the original forged images $I_{forgery}$ to synthesize pseudo labels $I_{p}$. $I_\triangle$, $I_{anti}$, and $I_{p}$ are used to calculate $Loss_{self}$ via Eq.~\ref{loss_self}. The anti-forensic images are fed into the supervisor, and we get the predicted masks $M_{pred}$. We compute $Loss_{adv}$ (Eq.~\ref{loss_adv}) using $M_{pred}$ and $M_{gt}$. The two losses are combined to obtain $Loss_{Concealer}$, which is defined as
\begin{equation}
	\label{loss_c}
	Loss_{Concealer} = \alpha Loss_{self}+\beta Loss_{adv},
\end{equation}
where $\alpha$ and $\beta$ control the weights of the attack loss and the adversarial loss. Here, we set $\alpha=2$ and $\beta=3$ to balance the attack loss and adversarial loss, and it makes our concealer learns more to fool the forensic models and maintain the image quality. The parameters of the concealer is then back-propagated and updated by minimizing the $Loss_{Concealer}$.

In the final step of a single training iteration, we unfreeze the parameters of the supervisor and compute the loss of the supervisor $Loss_{Supervisor}$ between the predicted model $M_{pred}$ and the ground-truth model $M_{gt}$. This loss is calculated as:
\begin{equation}
	\label{loss_s}
	Loss_{Supervisor} = -\lambda Loss_{adv},
\end{equation}
where $\lambda$ means the weights of training the supervisor in this system. We set the hyperparameter $\lambda$ to 0.5 to prevent over-training. After the joint training process, SEAR achieves convergence and yields a powerful concealer.

\subsection{Model Inference}

After joint training, the concealer network can conceal tampering traces and deceive forensic models. Notably, during inference, the concealer only requires a single original forged image, despite the use of the supervisor network and ground-truth masks in SEAR's training process. Specifically, the original forged images $I_{forgery}$ are fed into the well-trained concealer network, and the perturbation maps $I_\triangle$ are generated and added into $I_{forgery}$ to yield the anti-forensic images $I_{anti}$. These anti-forensic images are expected to evade detection by forgery localization models.

\begin{table*}
	
	\centering
	\caption{The performance of the attack algorithms on four benchmark datasets under white-box setting.}
	\label{tab:white-box}
	\resizebox{0.7\linewidth}{!}{
		
		\begin{tabular}{@{}cccc|cc|cc|cc@{}}
			\toprule
			&                          & \multicolumn{2}{c|}{NIST}                                 & \multicolumn{2}{c|}{Columbia}                                     & \multicolumn{2}{c|}{Coverage}                             & \multicolumn{2}{c}{CASIA}                                \\ \cmidrule(l){3-10} 
			\multirow{-2}{*}{Forensic Model} & \multirow{-2}{*}{Attack} & \makecell{Attack\\Rate} & \makecell{$F_1^R$ } & \makecell{Attack\\Rate} & \makecell{$F_1^R$ }         & \makecell{Attack\\Rate} & \makecell{$F_1^R$ }&\makecell{Attack\\Rate} & \makecell{$F_1^R$ } \\ \midrule
			& FGSM                     & 0.9154                  & 0.1003                         & 0.5511                  & 0.5377                                 & 0.7529                  & 0.2635                         & 0.8745                  & 0.1111                         \\
			& BIM                      & 0.9627                  & 0.4864                         & 0.8311                  & 0.7015                                 & 0.8677                  & 0.3239                         & 0.838                   & 0.8072                         \\
			& MIM                      & 0.9774                  & 0.4663                         & 0.781                   & 0.6742                                 & 0.8466                  & 0.3105                         & 0.8661                  & 0.7272                         \\
			& AdvGAN & 0.9239	& 0.0343 	& 0.5242	& 0.6771	& 0.5078	& 0.1691	& 0.6073	& 0.9359	\\
			\multirow{-5}{*}{Supervisor}      & SEAR                     & \textbf{0.9969}         & \textbf{0.0251}                & \textbf{0.9459}         & \textbf{0.0438}                        & \textbf{0.8962}         & \textbf{0.0601}                & \textbf{0.9953}         & \textbf{0.0028}                \\ \midrule
			& FGSM                     & 0.7344                  & 0.021                          & 0.5293                  & 0.5282                                 & 0.6617                  & 0.167                          & 0.872                   & 0.1533                         \\
			& BIM                      & 0.9861                  & 0.6869                         & 0.9384                  & 0.8191                                 & 0.9943                  & 0.8578                         & 0.9752                  & 0.9338                         \\
			& MIM                      & 0.9836                  & 0.6317                         & 0.9219                  & 0.8518                                 & 0.9904                  & 0.8787                         & 0.9625                  & 0.9262                         \\
			& AdvGAN & 0.8382	& 0.1856 	& 0.8517	& 0.7489	& 0.5872	& 0.6443	& 0.5523	& 0.7718	\\
			\multirow{-5}{*}{Mantra-Net}      & SEAR                     & 0.9368                  & \textbf{0.0165}                & 0.8319                  & {\textbf{0.1549}} & 0.7796                  & \textbf{0.0277}                & \textbf{0.9913}         & \textbf{0.0139}                \\ \midrule
			& FGSM                     & 0.7485                  & 0.0725                         & 0.5135                  & 0.7847                                 & 0.56                    & 0.8604                         & 0.905                   & 0.2927                         \\
			& BIM                      & 0.94                    & 0.5836                         & 0.8861                  & 0.7843                                 & 0.8455                  & 0.8913                         & 0.7466                  & 0.9467                         \\
			& MIM                      & 0.9409                  & 0.5339                         & 0.7079                  & 0.8565                                 & 0.6439                  & 0.878                          & 0.7531                  & 0.9359                         \\
			& AdvGAN & 0.862	& 0.0218 	& 0.5161	& 0.8063	& 0.5568	& 0.8521	& 0.8076	& 0.0586	\\
			\multirow{-5}{*}{SPAN}            & SEAR                     & 0.8702                  & \textbf{0.0134}                & \textbf{0.9356}         & \textbf{0.0017}                        & \textbf{0.8697}         & \textbf{0.0072}                & \textbf{0.9741}         & \textbf{0.0074}                \\ \bottomrule
		\end{tabular}
	}
\end{table*}

\section{Experiments}
\subsection{Setup}

\subsubsection{Datasets}
We trained and evaluated our SEAR system on several benchmark datasets, which contain three types of forgery techniques, namely splicing, copy-move, and removal. Specifically, we used the following datasets for training and testing:
\begin{itemize}
	\item {\bf NIST} 2016~\cite{guan_wacvw_2019} is a practical dataset since it involves three forgery techniques, containing 564 samples. 
	\item {\bf Columbia}~\cite{ng_cucdl_2009} focuses on spliced images with 180 samples.
	\item {\bf COVERAGE}~\cite{wen_icip_2016} focuses on copy-moved images with 100 samples.
	\item {\bf CASIA}~\cite{dong_csicsip_2013} contains spliced and copy-moved images of various objects with 6044 samples, including 921 samples of CASIA v1 and 5123 samples of CASIA v2.
	\item {\bf IMD}~\cite{novozamsky_wcavw_2020} is a large-scale and diverse dataset with various of artifacts, and consists of 2009 images.
\end{itemize}
To train our SEAR system, we divided each dataset into 75\% training and 25\% testing sets, following the same setting as in~\cite{zhuo_tifs_2022}. The selected datasets are sufficiently diverse to demonstrate the capabilities of anti-forensics. For convenience of training, we do not adopt the large datasets, like DEFACTO~\cite{mahfoudi_eusipco_2019} and PS-dataset~\cite{zhuang_tifs_2021} that appeared in SATFL~\cite{zhuo_tifs_2022}. Nevertheless, we employ IMD for only testing under black-box setting. We trained our system using input images resized to $512\times512$ on 8 Tesla P100 GPUs, with the ADAM solver at a learning rate of 0.0002.

\subsubsection{Metrics}

The pixel-level $F_1$ score is a commonly used evaluation metric in the literature to assess the performance of forensic models in binary classification tasks. The $F_1$ score measures the harmonic mean of precision and recall for each pixel, where values closer to 1 indicate better detection performance. The score ranges between 0 and 1, with higher values indicating better accuracy of the forensic model.

The primary focus of the Pixel-level $F_1$ score is to assess the accuracy of the forensic model. However, anti-forensic techniques aim to deceive forensic analysis and conceal evidence, necessitating the development of a new metric to evaluate the effectiveness of these techniques. Therefore, we propose to measure the attack performance against the forensic model using the following attack rate:

\begin{equation}
	Attack\ Rate = \frac{F^{ori}_1 - F^{anti}_1}{F^{ori}_1},
\end{equation}
where $F^{ori}_1$ denotes the $F_1$ score of detection on the original forged images, and $F^{anti}_1$ means the $F_1$ score of detection on the anti-forensic images. The attack rate quantifies the extent of degradation in $F_1$ score of a forensic model after an anti-forensic attack. A higher attack rate indicates a more effective attack algorithm. The attack rate offers a more precise and dependable measure of the effectiveness of anti-forensic techniques. Moreover, to assess the image quality of the anti-forensic images generated by our approach, we employ the peak signal to noise ratio (PSNR) and structural similarity index (SSIM) as evaluation metrics. Higher PSNR and SSIM values indicate superior image quality. \footnote{For implementation details, the source code is available at https://github.com/tansq/SEAR.}

Furthermore, we introduce $F_1\ of\ Reverse$ ($F_1^R$) to calculate $F_1$ score for the reversed masks of the predicted masks. Since tampered pixels' values in $M_{pred}$ are one while the non-tampered pixels' values are zero, we calculate the reversed masks as 
\begin{equation}
	M_{reversed} = M_{1}- M_{pred},
\end{equation}
where $M_{1}$ is a mask that has the same size as $M_{pred}$, and each pixel's value is set to 1.

The reversed masks denote that the value of the predicted non-tampered pixels in $M_{reversed}$ is one, whereas that of the predicted tampered pixels is zero. Based on our observation, we figure out that the zero-one reversal phenomenon occurs when the reversed masks of predicted masks for the anti-forensic images resemble the ground-truth masks. The $F_1^R$ metric quantifies the severity of the reversed phenomenon, where a smaller value indicates a lower severity.

\subsubsection{Forensic Models} 

In our experiments, we evaluated the performance of three different forgery localization models: our supervisor, Mantra-Net, and SPAN. Since we focus on pixel-level anti-forensics, the selected methods are only focus on pixel-level forensics. Our supervisor employs the same structure to SATFL~\cite{zhuo_tifs_2022}, and we also implemented two other notable forgery localization models: Mantra-Net~\cite{wu_cvpr_2019} and SPAN~\cite{hu_eccv_2020}. Mantra-Net is a deep network based on LSTM to extract local manipulations, while SPAN uses a pyramid attention structure to detect tampered regions. The forensic models have been well-trained before the attacks.

\subsubsection{Other Attack Methods}

We compared well-known adversarial attack algorithms, including FGSM~\cite{goodfellow_arxiv_2014}, BIM~\cite{kurakin_arxiv_2016}, MIM~\cite{dong_cvpr_2018} and AdvGAN~\cite{xiao_arxiv_2018}. 

We selected the aforementioned techniques as adversaries for three main reasons. Firstly, we aim to conduct the comprehensive experiments with both non-learnable and learnable attack methods. Secondly, they demonstrate excellent attack performance, which is crucial for ensuring the reliability and accuracy of the results. Thirdly, they exhibit relatively fast inference speed, which is essential for practical applications. 

\subsubsection{Attack Situations}
In practice, attack algorithms may encounter three different scenarios: white-box, black-box, and defense settings. In our paper, we evaluate our approach in all three of these situations.

\begin{itemize}
	\item {\bf White-box setting:} the attack algorithms have access to the forensic models. Similar to adversarial attacks, anti-forensic methods can obtain the gradients, parameters, and outputs of the forensic models.
    \item {\bf Black-box setting:} the attack algorithms have no access to the forensic models. This means that the attack methods have no prior knowledge of the model's parameters or training data. To simulate this setting and compare our method with non-learnable attacks, we adopt the same approach as~\cite{xiao_arxiv_2018}. First, we create distilled models with small structures that act as local forensic models since non-learnable attacks need a target model. The local forensic models are trained on a subset of instances that are disjoint from the training data. The local forensic models serve as the targets of the attack methods. It's worth noting that the local forensic models have different structures, gradients, and parameters compared to the forensic models. Thus, in this setting, the attack methods remain unaware of the forensic models' parameters and gradients. Furthermore, regarding the comparison with learnable attacks, we set a purely black-box setting without local forensic model.
	
    \item {\bf Defense setting:} the forensic models have learned to defend against attack algorithms. This means that the forensic models have access to the attack methods and are able to defend against them. One powerful defensive approach is retraining defense~\cite{xu_ijac_2020}, which involves retraining the forensic models with the attacked images as new training data. This allows the forensic models to learn the features of the perturbation maps and accurately localize the attacked regions. In this setting, we retrain the forensic models using the anti-forensic images of the aforementioned attack methods, and subsequently, the attack methods are applied to the retrained forensic models in the white-box setting.

\end{itemize}

\subsection{Evaluation}

Here, we have compared the results of our SEAR with those of other well-known adversarial attack algorithms on the NIST, the CASIA, the COVERAGE, and the Columbia dataset. Due to the space limitation, we visualize more examples in the supplementary materials.

\subsubsection{White-box Setting}

In the white-box setting, the attack algorithms (FGSM, BIM, MIM, AdvGAN and SEAR) are used to perform anti-forensic attacks against the forensic models, namely SATFL, Mantra-Net, and SPAN. 

We argue that the attack algorithms tend to focus excessively on maximizing the loss of the forensic models, resulting in the widespread occurrence of the zero-one reversal phenomenon. The adversarial attack aims to maximize the loss between the ground-truth and predicted masks. The zero-one reversal seems to be the maximum loss of them. To account for this, we use $F_1^R$ for evaluation in this setting.

\noindent
\textbf{Quantitative Results.}
Table~\ref{tab:white-box} presents the detailed comparison results of SEAR with other popular adversarial attack algorithms in the white-box setting. As shown, SEAR achieves the best performance for the attack rate without the zero-one reversal phenomenon. In particular, regarding the zero-one reversal phenomenon, for attacking three forensic models on four datasets, FGSM gains $F_1^R$ values ranging from 0.1003 to 0.8604, BIM ranges from 0.3239 to 0.9467, MIM ranges from 0.3105 to 0.9359 and AdvGAN obtains from 0.0218 to 0.9359. It demonstrates that even after being attacked by FGSM, BIM, MIM, and AdvGAN the forensic models still achieve high forensic scores of the reverse masks, around 0.9, indicating that the forensic models label most areas of pristine regions as the tampered ones but label the most tampered regions as the pristine ones. In contrast, SEAR achieves a low $F_1^R$ score, ranging from only 0.0028 to 0.1549, most of which are less than 0.01. It indicates that the reversed masks of predicted masks for the anti-forensic images by our SEAR are essentially unaffiliated with the ground-truth masks, while other methods' predicted masks are quite affiliated with the ground-truth masks. Therefore, the typical adversarial attack algorithms, especially BIM, suffer from the zero-one reversal phenomenon while our SEAR has overcome this challenge.

\begin{figure*}
	\centering
	
	\includegraphics[width=1.0\linewidth]{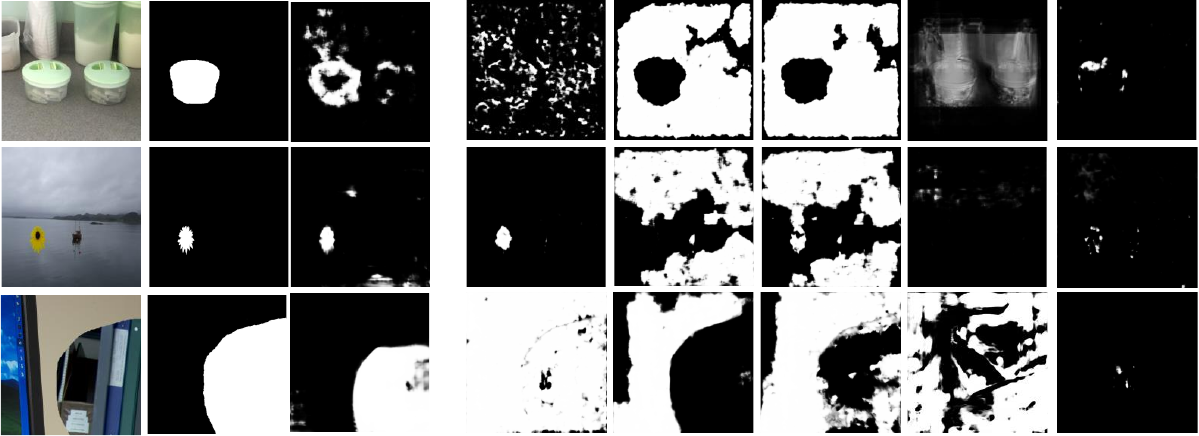}
	\caption{The comparison results of the predicted masks of the anti-forensic images by the different attacked methods in the white-box setting. The first column is the original forged images, the second column is the ground-truth masks, the third column is the predicted masks of the original forged images, and from the fourth to the seventh column are the predicted masks of the anti-forensic images generated by FGSM, BIM, MIM, AdvGAN and our SEAR.} 
	\label{visual_white}
\end{figure*}

\begin{table}
	
	\centering
	\caption{The visual quality of the attack algorithms on four benchmark datasets under white-box setting. The higher score means better visual performance.}
	\label{tab:quality}
	\resizebox{1\linewidth}{!}{
		\begin{tabular}{@{}cccc|cc|cc|cc@{}}
			\toprule
											 &                          & \multicolumn{2}{c}{NIST}         & \multicolumn{2}{c}{Columbia}                                                   & \multicolumn{2}{c}{Coverage}     & \multicolumn{2}{c}{CASIA}        \\ \cmidrule(l){3-10} 
			\multirow{-2}{*}{Forensic Model} & \multirow{-2}{*}{Attack} & SSIM            & PSNR           & SSIM                                   & PSNR                                  & SSIM            & PSNR           & SSIM            & PSNR           \\ \midrule
											 & FGSM                     & 0.9374          & 38.61          & 0.9589                                 & 38.62                                 & 0.9653          & 34.63          & 0.9314          & 31.62          \\
											 & BIM                      & 0.9897          & 46.41          & 0.9699                                 & 37.72                                 & 0.9621          & 32.5           & 0.945           & 35.94          \\
											 & MIM                      & 0.9726          & 41.91          & 0.9794                                 & 36.56                                 & 0.9764          & 35.44          & 0.9342          & 37.11          \\
											 & AdvGAN	& 0.9241 &	40.81	&	 0.9147	&	36.22	&	{0.9987}	&	{59.84} &   {0.978}   &	36.07  \\
			\multirow{-5}{*}{Supervisor}    & SEAR                     & \textbf{0.9982} & \textbf{53.96} & \textbf{0.9816}                        & \textbf{38.96}                        & {0.9794} & {36.29}         & {0.9554} & \textbf{38.28}          \\\midrule
			
											 & FGSM                     & 0.9077          & 38.61          & 0.9118                                 & 38.62                                 & 0.9299          & 38.63          & 0.9682          & 38.62          \\
											 & BIM                      & 0.9568          & 40.21          & 0.9581                                 & 37.63                                 & 0.9411          & 38.87          & 0.9659          & 37.42          \\
											 & MIM                      & 0.9553          & 41.95          & 0.959                                  & 38.03                                 & 0.9672          & 42.11          & 0.9845          & 36.78          \\
											 & AdvGAN	& 0.8667	&	36.34 &	{0.9698}	&	{43.4}	&	0.9597	&	40.81 &   0.9641  &	36.23   \\
			\multirow{-5}{*}{Mantra-Net}     & SEAR                     & \textbf{0.9801} & \textbf{42.36} & {{0.9631}} & {{39.26}} & \textbf{0.9695}          & \textbf{43.09} & \textbf{0.9864} & \textbf{39.67}          \\\midrule
											 & FGSM                     & 0.9092          & 35.61          & 0.9142                                 & 38.73                                 & 0.9301          & 35.63          & 0.9684          & 38.63          \\
											 & BIM                      & 0.9454          & 36.51          & 0.9884                                 & 37.79                                 & 0.9802          & 36.91          & 0.996           & 38.16          \\
											 & MIM                      & 0.934           & 32.7           & 0.9573                                 & 38.99                                 & 0.9646          & 37.66          & 0.9828          & 39.55          \\
											 & AdvGAN	& 0.9294 &	{40.36}	 & 0.972 &	{43.64} &	{0.9976}	&	{52.97}	&   0.9809  &	{47.88}   \\
			\multirow{-5}{*}{SPAN}           & SEAR                     & \textbf{0.9425} & {38.5}  & \textbf{0.9896}                        & {39.06}                                 & {0.9836} & {38.38}          & \textbf{0.9973} & {45.52} \\ \bottomrule 
			\end{tabular}
		
	}
\end{table}

With respect to the attack rate, our SEAR algorithm outperforms other adversarial attack algorithms on all four datasets, achieving a range of 0.8962 to 0.9969 attack rates for the supervisor forensic model. This represents a significant improvement over other methods, with a margin of 0.0198 to 0.3948. When compared to FGSM, SEAR also achieves better attack rates against Mantra-Net and SPAN, with a margin of 0.0691 to 0.4221. Additionally, compared to learnable AdvGAN, SEAR outperforms by 0.4295 at best. Moreover, BIM and MIM encounter serious zero-one reversal when attacking the Mantra-Net and SPAN models.

One possible reason for such experimental results is that other methods attack the forensic models by maximizing the losses of the forensic model. They perturb the original forged images globally to fool the forensic models to detect the pristine regions as the tampered ones and the tampered regions as pristine ones. In contrast to other methods that perturb the original forged images globally to deceive the forensic models, self-supervised learning restricts the perturbation into the tampered regions, allowing for the adaptive elimination of tampering traces. We attribute SEAR's superior performance to the flexibility of self-supervised learning, which guides the model to smooth the tampering traces effectively, as demonstrated in Sect.~\ref{sec:how}. This capability plays a crucial role in overcoming the zero-one reversal phenomenon.

\begin{figure*}[t]
	\centering
	
	\includegraphics[width=1\linewidth]{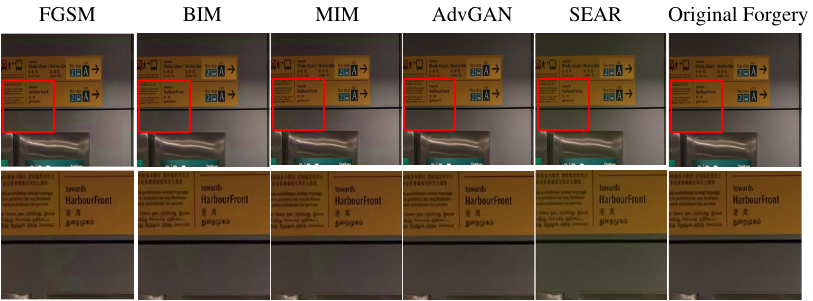}
	\caption{The comparison results in the visual quality of the anti-forensic images generated by different attack methods. The first row contains the anti-forensic images and the original forged image, and the second row enlarges the regions in the red boxes.}
	\label{visual_quality}
\end{figure*}

We have also evaluated the visual quality of the attacked images generated by the four methods in terms of SSIM and PSNR. Table~\ref{tab:quality} shows the results, and it can be observed that SEAR outperforms the other methods in terms of visual quality. Specifically, SEAR achieves SSIM values ranging from 0.9425 to 0.9982 and PSNR values ranging from 38.28 to 53.96 when attacking the three forensic models on four datasets. Fig.~\ref{visual_quality} further shows that SEAR generates smoother results, while  the attacked images generated by other methods contain significant artifacts. These results suggest that SEAR is capable of generating high-quality anti-forensic images with less damage to the original forged images.

\noindent
\textbf{Qualitative Results.}
Furthermore, we have illustrated more comparison examples. As shown in Fig.~\ref{visual_white}, in white-box setting, BIM and MIM have led to severe reversed phenomenon where most of pristine parts are predicted as tamped regions. In addition, FGSM and AdvGAN have slight attacking effects on misleading the forensic methods, implying suboptimal attack performance. 
In contrast, SEAR have fooled the forensic model to detect the tampered regions as the pristine ones without the reversed phenomenon.

\subsubsection{Black-box Setting}

\begin{table}

		\centering
		\caption{The performance of the attack algorithms on four benchmark datasets under black-box setting. }
		\label{black_box}
		\resizebox{0.95\linewidth}{!}{
			\begin{tabular}{@{}cccccc@{}}
				\toprule
				\multirow{2}{*}{Forensic Model} & \multirow{2}{*}{Attack} & \multicolumn{4}{c}{Attack Rate}                           \\ \cmidrule(l){3-6} 
				&                         & NIST            & Columbia        & Coverage        & Casia           \\ \midrule
				\multirow{5}{*}{Mantra-Net}      & FGSM                    & 0.7855          & 0.5995          & 0.5995          & 0.6857          \\
				& BIM                     & 0.5728          & 0.2722          & 0.5737          & 0.4899          \\
				& MIM                     & 0.8515          & 0.4749          & 0.5581          & 0.6652          \\
				& AdvGAN	&	0.7946	&	0.7016	&	0.1124	&	0.6123	\\ 
				& SEAR                    & \textbf{0.9557} & \textbf{0.9438} & \textbf{0.8658} & \textbf{0.9343} \\ \midrule
				\multirow{5}{*}{SPAN}            & FGSM                    & 0.6159          & 0.5161          & 0.5418          & 0.9291          \\
				& BIM                     & 0.3535          & 0.4546          & 0.5213          & 0.7463          \\
				& MIM                     & 0.5967          & 0.5011          & 0.522           & 0.8555          \\
				& AdvGAN	&	0.7911	&	0.6045	&	0.1857	&	0.7545	\\ 
				& SEAR                    & {0.7192} & {0.5809} & \textbf{0.8703} & \textbf{0.9668} \\ \bottomrule
			\end{tabular}
		}
	
\end{table}

\begin{figure*}
	\centering
	
	\includegraphics[width=1.0\linewidth]{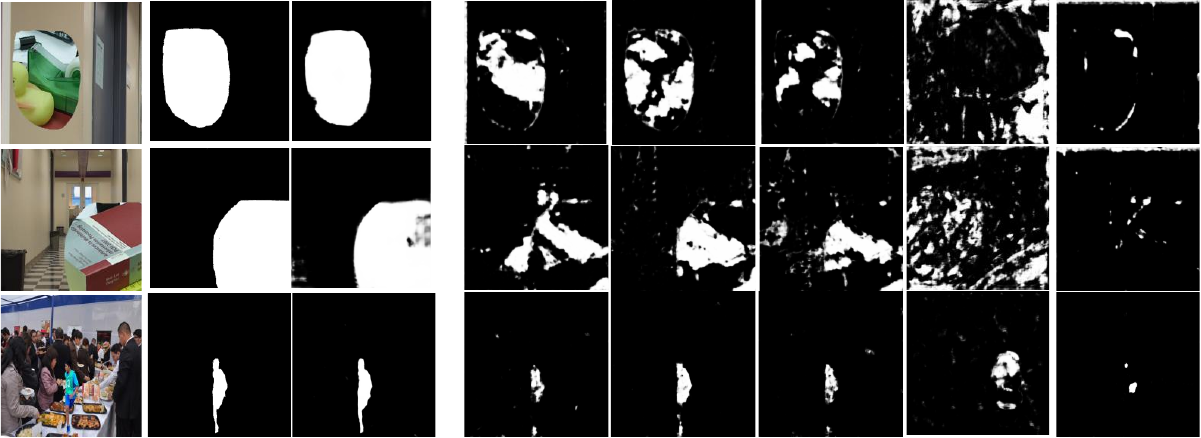}
	\caption{The comparison results of the predicted masks of the anti-forensic images by the different attacked methods in the black-box setting. The first column is the original forged images, the second column is the ground-truth masks, the third column is the predicted masks of the original forged images, and from the fourth to the seventh column are the predicted masks of the anti-forensic images generated by FGSM, BIM, MIM, AdvGAN and our SEAR.}
	\label{visual_black}
\end{figure*}

To evaluate the generalization ability of the attack algorithms, we have further conducted block-box experiments on four datasets. Specifically, the distilled models of Mantra-Net and SPAN were used as the substitute models for the real forensic models. The attack approaches generate the anti-forensic images against the distilled models. In this setting, the real forensic models were unknown to the attack algorithms, which was a challenging task to evaluate the generalization ability. In purely black-box setting, the well-trained AdvGAN and SEAR under white-box setting using NIST dataset are utilize to predict the adversarial example maps of IMD dataset.

\noindent
\textbf{Quantitative Results.}
The results with the distilled model are reported in Table~\ref{black_box}. Please note that the zero-one reversal phenomenon has been alleviated in the black-box setting since the targeted gradients of the forensic models are unseen to the attack methods.~\footnote{The examples of the black-box setting are shown in supplementary materials.} Consequently, only the attack rate is reported in Table~\ref{black_box}. Our SEAR again achieves the best performance on four benchmark datasets when attacking two forensic models, respectively. Specifically, for attacking Mantra-Net, our SEAR achieves an attack rate ranging from 0.8658 to 0.9557 and outperforms other attack methods by a significant margin of 0.1042 to 0.7534 across the four datasets. Regarding attacking SPAN, our SEAR obtains an attack rate ranging from 0.7192 to 0.9668 and surpasses other attack methods among the four datasets, representing an improvement of 0.0648 to 0.6846 over other attacks. The results also show that the adversarial attack methods are model-specific, and their effectiveness can be limited when facing other forensic models. In contrast, our SEAR benefits from the inclusion of the supervisor and adversarial training, and it exhibits significantly improved generalization ability.

As shown in Table~\ref{black_box2}, in purely black-box setting, our SEAR achieves attack rate that is 160\% of AdvGAN on IMD dataset.
Both AdvGAN and our SEAR only generate adversarial example maps of the corresponding images in IMD dataset without any extra training. The results show that benefits from self-supervised, SEAR exhibits more powerful generalization performance than AdvGAN which relies solely on adversarial training.

\begin{table}

		\centering
		\caption{The performance of the attack algorithms on four benchmark datasets under black-box setting on IMD dataset. }
		\label{black_box2}
		\resizebox{0.65\linewidth}{!}{
			\begin{tabular}{@{}ccc@{}}
				\toprule
				\multirow{2}{*}{Forensic Model} & \multirow{2}{*}{Attack} & \multicolumn{1}{c}{Attack Rate}                           \\ \cmidrule(l){3-3}  & & IMD           \\ \midrule
				\multirow{2}{*}{SATFL}      
			
				& AdvGAN	&	0.3059	\\ 
				& SEAR                    & \textbf{0.8019} \\  \bottomrule
			\end{tabular}
		}
	
\end{table}

\noindent
\textbf{Qualitative Results.}
Fig.~\ref{visual_black} demonstrates that in black-box setting, the forensic models can still accurately localize the tampered regions when attacked by FGSM, BIM MIM, and AdvGAN. This suggests these methods fail to fool the forensic models under black-box setting. Whereas our SEAR successfully deceives the forensic models and conceals the tampered traces, representing that the forensic models ignore the most of tampered regions.  

\subsubsection{Retrained Defense Setting}

\begin{table}

	\centering
	\caption{The performance of the attack algorithms on NIST dataset under retrained defense setting. Before retraining, the supervisor's detection performance $F_1$ is 0.846, Mantra-Net's is 0.7941, and SPAN's is 0.7813. }
	\label{retrain}
	\resizebox{0.8\linewidth}{!}{
		\begin{tabular}{@{}cccc@{}}
			\toprule
			Forensics Model        & Attack & \makecell{$F_1$\\after Retraining} 
			   & \makecell{Attack\\Rate} \\ \midrule
			\multirow{4}{*}{Supervisor} & FGSM   & 0.8919                          & 0.036           \\
			& BIM    & 0.8805           & 0.3261          \\
			& MIM    & 0.88                            & 0.343           \\
			& SEAR   & 0.8682                & \textbf{0.9619} \\ \midrule
			\multirow{4}{*}{Mantra-Net} & FGSM   & 0.8239                          & 0.1832          \\
			& BIM    & 0.8232                     & 0.7834          \\
			& MIM    & 0.7828                          & 0.7917          \\
			& SEAR   & 0.8124                          &\textbf{ 0.8279}          \\ \midrule
			\multirow{4}{*}{SPAN}       & FGSM   & 0.8008                          & 0.1526          \\
			& BIM    & 0.8263                    & {0.5179}          \\
			& MIM    & 0.8126                         & 0.3864          \\
			& SEAR   & 0.8047                          & 0.4571          \\ \bottomrule
		\end{tabular}
	}
	
\end{table}

In addition to the aforementioned situations, forensic models also proactively employ defensive techniques to prevent attacks, as previously reported by Xu et al.~\cite{xu_ijac_2020}. We retrained the forensic models using the attacked samples. During retraining, the forensic models were able to learn the attacked features and render the attacks useless. The experiments were conducted on the NIST dataset, and the forensic models were retrained using the anti-forensic images generated by each attack method, respectively. Following the retraining process, the attack methods were used again to generate new anti-forensic images under the white-box setting.

\noindent
\textbf{Quantitative Results.}
The results are presented in Table~\ref{retrain}. The results indicate that the forensic models have become more robust after being retrained with the generated anti-forensic images, making it more difficult to deceive them. Nevertheless, our SEAR approach still achieves superior performance when attacking three of the forensic models. Notably, SEAR attains an attack rate of 0.9619 when targeting the supervisor, surpassing FGSM by 0.9259. When attacking Mantra-Net and SPAN, SEAR surpasses FGSM, BIM, and MIM by ranging from 0.0707 to 0.6447 in terms of attack rate. These results demonstrate that SEAR is a flexible model to fight against the defensive method. By employing adversarial learning, the concealer can attack those versatile forensic models.

\begin{figure*}	  
	\includegraphics[width=1.0\linewidth]{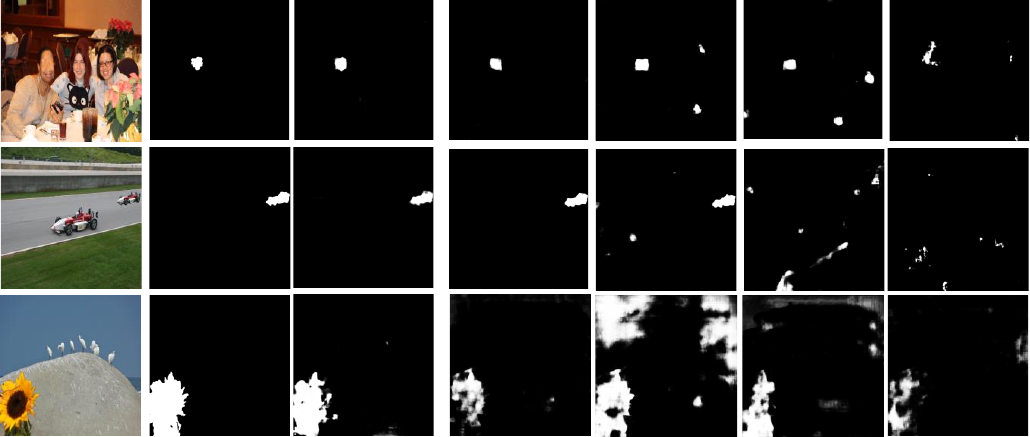}
	\caption{The comparison results of the predicted masks of the anti-forensic images by the different attacked methods in the retrained defense setting. The first column is the original forged images, the second column is the ground-truth masks, the third column is the predicted masks of the original forged images, and from the fourth to the seventh column are the predicted masks of the anti-forensic images generated by FGSM, BIM, MIM and our SEAR.}
	\label{visual_adv}
\end{figure*}

\noindent
\textbf{Qualitative Results.}
As illustrated in Fig.~\ref{visual_adv}, in the retrained defense setting, the forensic models can still precisely localize the tampered regions when attacked by FGSM, BIM and MIM. This indicates the inferior attack capability of these methods against the retrained forensic models. On the contrary, under the attack of our proposed SEAR approach, the retrained forensic models fail to output correct predictions and wrongly recognizes the tampered areas as authentic.

\subsubsection{Inference Speed}

\begin{table}
	\centering
	\caption{Inference time (seconds) of different attack algorithms for a image. }
	\label{infer_time}
	\resizebox{0.48\textwidth}{!}{
	\begin{tabular}{@{}cccccl@{}}
		\toprule
		Attack & NIST   & Coverage & Columbia & Casia  & \multicolumn{1}{c}{Average} \\ \midrule
		FGSM   & 0.2021 & 0.2019   & 0.207    & 0.2055 & 0.2041                    \\
		BIM    & 0.9419 & 0.9438   & 0.9513   & 0.9421 & 0.9447                    \\
		MIM    & 1.1316 & 1.1375   & 1.1476   & 1.1468 & 1.1408                    \\
		AdvGAN	& 0.1872 & 0.1923	& 0.1866	& 0.171 & 0.1843					\\
		\midrule
		SEAR   & \textbf{0.1592} & \textbf{0.1727}   & \textbf{0.1614 }  &\textbf{ 0.1523} & \textbf{{0.1614}}             \\ \bottomrule
	\end{tabular}
 }

\end{table}

We have calculated the inference time of each attack method on four datasets to evaluate the speed of the different attack methods on a single Tesla P100 GPU. The inference time is computed by averaging the total time for attacking the testing sets of four datasets, respectively. The inference time represents the time that the attack method needs to generate an anti-forensic image with a resolution of $512\times 512$. We have also averaged the inference times in Table~\ref{infer_time}. As shown, our method has the fastest inference speed among FGSM, BIM, MIM and AdvGAN. On average, our approach only needs 0.1614 seconds to generate an anti-forensic image, which is almost $\frac{1}{10}$ of the inference time of MIM. This indicates that our concealer is practical due to its lightweight structure.

\subsection{Ablation Study}

We have conducted ablation studies on the proposed SEAR using the NIST dataset, which is a well-established benchmark for evaluating image anti-forensic techniques and comprises three commonly encountered content-based manipulations. The purpose of these studies is twofold: to examine the efficacy of our proposed self-supervised learning approach, and to evaluate the contribution of convolutional blocks to the performance of our model.

\subsubsection{Effect of the proposed self-supervised learning}

Our study aims to investigate the benefits of incorporating self-supervised learning for anti-forensic techniques. In our ablative experiments, we first presented the results of the baseline, which is our method without self-supervised learning, referred to as ``w/o Self-Supervised''. We then reported the performance of the proposed SEAR with self-supervised learning, denoted as ``w/ Self-Supervised''.

\begin{figure}
	\centering
	
	\includegraphics[width=1\linewidth]{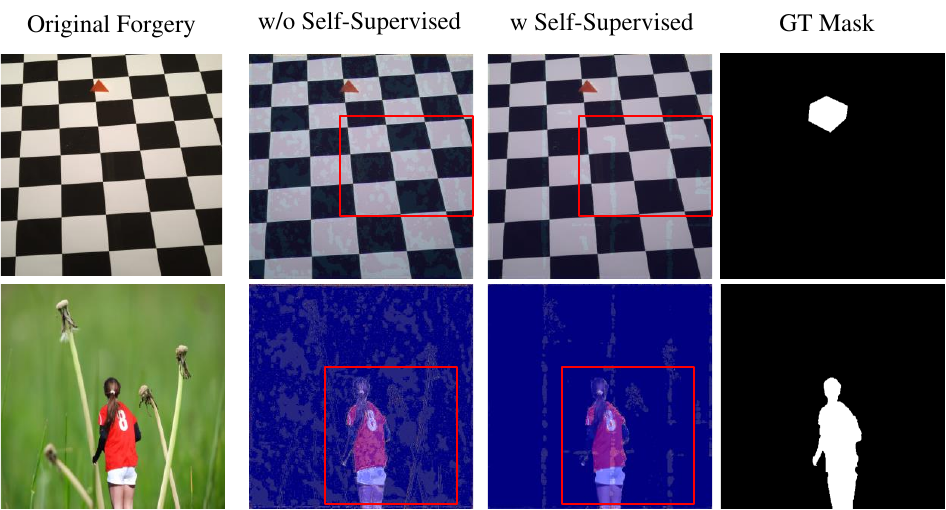}
	\caption{The comparison samples between training with self-supervised learning (w/ Self-Supervised) and training without self-supervised learning (w/o Self-Supervised). The heatmaps of the perturbation maps are super-imposed on the corresponding anti-forensic images. The red boxes highlight the differences between the perturbation maps.}
	\label{abl_visual_ss}
\end{figure}

The visualization analysis, presented in Fig.~\ref{abl_visual_ss}, demonstrates a significant difference in the perturbation maps between the models trained with and without self-supervised learning. Specifically, the perturbation map without self-supervised learning exhibits massive, high-intensity perturbations that are scattered throughout the original forged images. In contrast, the perturbation map with self-supervised learning is concentrated in the tampered regions, with fewer perturbations in pristine areas. For example, in the first line of Fig.~\ref{abl_visual_ss}, the perturbation map with self-supervised learning shows minimal perturbation in the pristine regions (indicated by red boxes). Similarly, in the second line of Fig.~\ref{abl_visual_ss}, the red boxes highlight that the perturbation with self-supervised learning is concentrated in the tampered area (the person) and ignores the background. These results suggest that self-supervised learning guides the concealer to focus on perturbing the tampered regions while ignoring the pristine areas, resulting in high-quality anti-forensic images.

\begin{table}[t]
	\centering
	\caption{The ablation study for self-supervised learning. The results are reported in (tampered regions' modifications/ pristine regions' modifications).}
	\label{tab:abl_num_ss}
	\resizebox{0.4\textwidth}{!}{
		\begin{tabular}{@{}cccc@{}}
			\toprule
			Attack & Mean Values   & Variance Values \\ \midrule
			w/o Self-Supervised   & 2.95/3.04 & 17.2/14.47                      \\
			    
			w/ Self-Supervised   & \textbf{2.38/2.18} & \textbf{15.4/11.11}      \\ \bottomrule
		\end{tabular}
	}
\end{table}

We have also conducted a statistical analysis on the perturbation maps generated by SEAR with and without self-supervised learning on the NIST dataset. This analysis calculated the mean values in the tampered regions and the variance values in the pristine areas of the perturbation maps. As shown in Table~\ref{tab:abl_num_ss}, the mean value in the tampered regions and the variance value in non-tampered regions of the perturbation with self-supervised learning are lower than those without self-supervised learning. We attribute this to the perturbation restraint introduced by self-supervised learning. These results confirm that self-supervised learning helps restrict perturbation in high-frequency regions and preserve image quality.

\begin{table}[t]
	\centering
	\caption{The ablation study for self-supervised learning on attack performance.}
	\label{tab:abl_ss}
	\resizebox{0.7\linewidth}{!}{
		\begin{tabular}{@{}cccc@{}}
			\toprule
			Attack & Attack Rate   & SSIM & PSNR \\ \midrule
			w/o Self-Supervised   & 0.7224 & 0.9796   & 41.88                     \\
			    
			w/ Self-Supervised   & \textbf{0.8414} & \textbf{0.9982}   & \textbf{53.96 }             \\ \bottomrule
		\end{tabular}
	}
\end{table}

To evaluate the effectiveness of self-supervised learning in anti-forensic images, we measured the attack rate and visual quality of images with and without self-supervised learning. As shown in Table~\ref{tab:abl_ss}, self-supervised learning improves the attack rate by a significant margin, with over 16\% improvement. Additionally, self-supervised learning benefits image quality, as SEAR with self-supervised learning achieves optimal performance in terms of SSIM and PSNR. Specifically, self-supervised learning improves PSNR by 29\% and SSIM by 2\%.

\begin{figure}[t]
	\centering
	\includegraphics[width=1\linewidth]{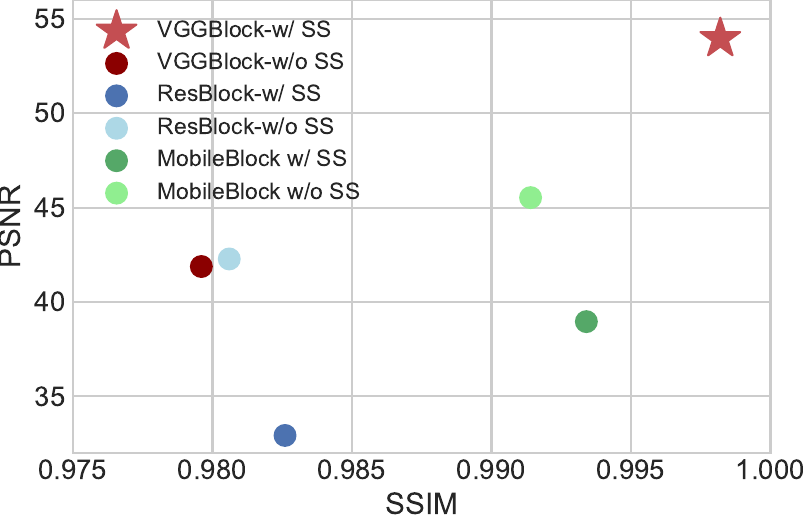}
	\caption{The visual quality of ablative study. }
	\label{abl_blocks}

	\centering

\end{figure}

\subsubsection{Effect of basic convolutional blocks}
\label{sec:abl-vgg}

In our concealer, we have utilized VGGBlock~\cite{simonyan_arxiv_2014} as the basic convolutional block to construct the network, which is a stable convolutional block that works well with the proposed self-supervised learning. The self-supervised learning introduces the tampered locations into the system, which requires a network with strong representations. While there are many advanced architectures of convolutional blocks, such as ResBlock~\cite{he_cvpr_2016} and MobileBlock~\cite{sandler_cvpr_2018}, we conducted ablative experiments to compare the basic convolutional blocks.

We built concealer networks using VGGBlock, ResBlock, and MobileBlock and trained them with or without self-supervised learning. The comparison results are presented in Fig.~\ref{abl_blocks}. It is evident that VGGBlock with self-supervised learning produces the most realistic anti-forensic images, achieving a PSNR score of over 50 and an SSIM score of 0.9982. Conversely, other blocks with self-supervised learning generated anti-forensic images with worse visual quality than those generated without self-supervised learning. It indicated that VGGBlock is a stable block with robust representations, making it suitable to work in tandem with self-supervised learning for better attack rates and visual quality.

	
		
		


\subsection{How our Concealer Works}
\label{sec:how}
\begin{figure*}
	\centering
	
	\includegraphics[width=0.9\linewidth]{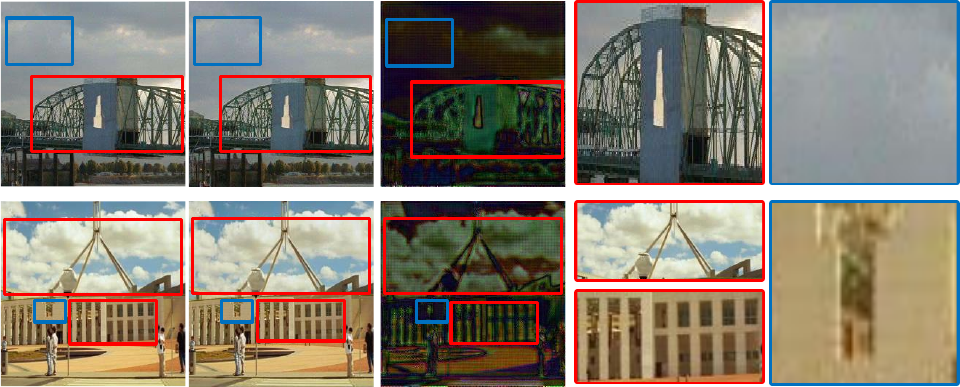}
	\caption{The residual images between anti-forensics images and original forgery images. The first column is the original forged images, the second is the anti-forensics images, and the third is the residual maps enlarged by 10$\times$ between the first and second columns. The red boxes show the regions of main modifications by our method, while the blue boxes show the areas of few modifications. The fourth and the fifth columns show the resized regions of red and blue boxes.}
	\label{analysis}
\end{figure*}

We have further explored how our concealer works in hiding the artifacts of tampering regions. We use the PIL package~\footnote{https://pypi.org/project/Pillow/} to calculate residual images, which reveal the difference between the original forgery and the anti-forensic images. Fig.~\ref{analysis} shows the residual images, with a 10-fold enhancement to improve the visual presentation,  where the light regions represent more differences compared to the dark regions. The amount of difference indicates the extent of modifications made by our method. We used red and blue boxes to highlight the areas that were highly and minimally modified, respectively, in the original tampered images by our method. Upon closer inspection of the red boxes, it was observed that our concealer mostly modified regions with complex textures, such as the bridge with a complicated structure. Our comparison results indicate that our concealer network smooths the images, concealing the high-frequency information, and blending the artifacts into the background.

\begin{figure}
	\centering
	\includegraphics[width=1\linewidth]{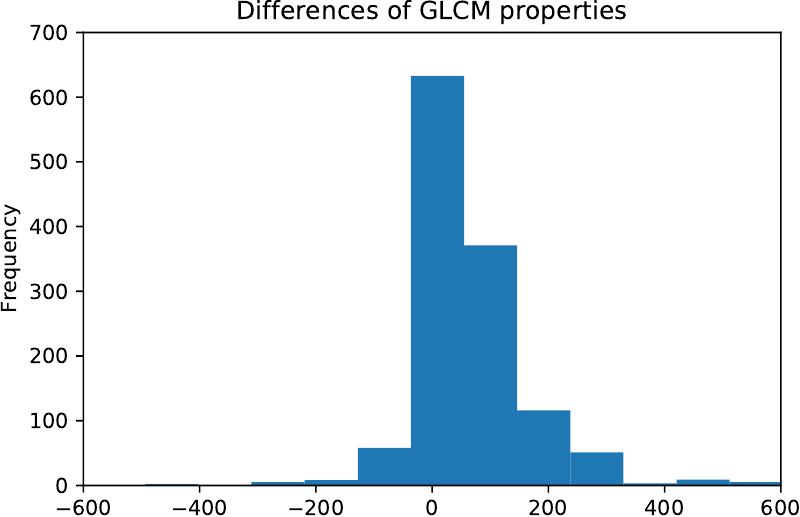}
	\caption{The differences of GLCM properties. The horizontal coordinate data is equal to $GLCM_{origin}-GLCM_{anti}$ }
	\label{hist} 
\end{figure}

We have also conducted a texture analysis on the generated images using the method proposed by Haralick~\cite{haralick_tsmc_1973}. Specifically, we utilized grey level co-occurrence matrices (GLCM), which are histograms of co-occurring grayscale values at a given offset over an image. GLCM properties reflect local grayscale variation in images, and higher properties indicate sharper image edges and higher frequency. To compare the texture of the original forged and anti-forensic images, we subtracted the GLCM properties of the anti-forensic images from those of the corresponding original forged images. This difference is denoted as $GLCM_{origin}-GLCM_{anti}$. A positive value indicates that the original forged image contains higher frequency information than the corresponding anti-forensic image. The texture analysis was based on the anti-forensic images generated by SEAR among four datasets. We have calculated $GLCM_{origin}-GLCM_{anti}$ for each image pair, and all of these values were counted into a histogram, as shown in Fig.~\ref{hist}. Most of the data points have positive values, indicating that the GLCM properties of the original forged images are statistically greater than those of the anti-forensic images. One possible explanation is that our concealer hides the artifacts by smoothing the regions of high-frequency information.

\begin{figure*}
	\centering
	
	\includegraphics[width=0.9\linewidth]{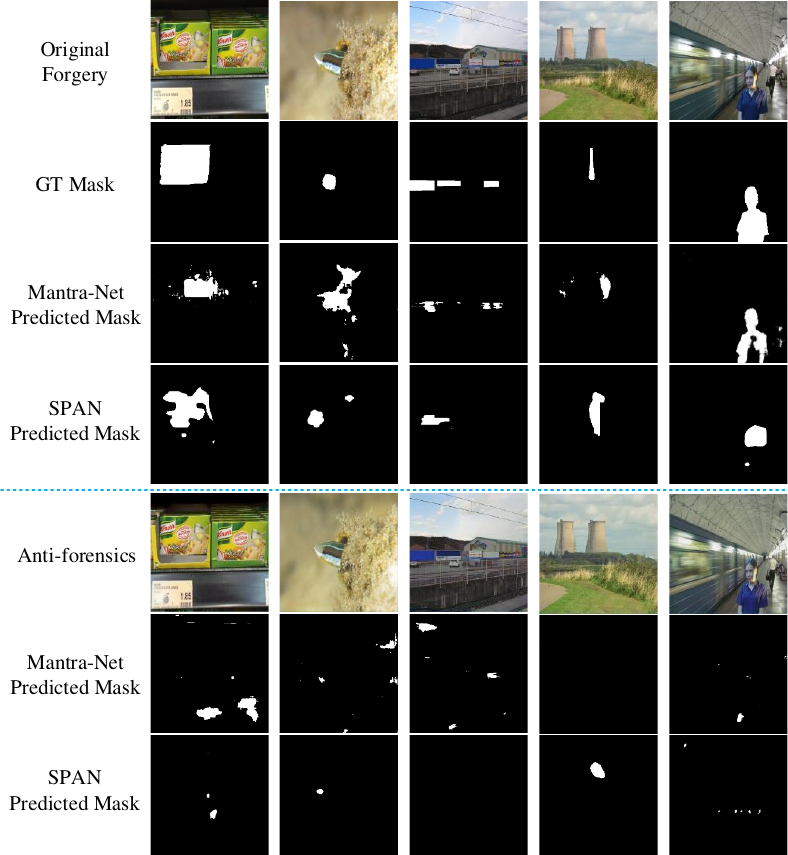}
	\caption{Performance of our system fighting against with the advanced forensic models.}
	\label{performance}
\end{figure*}

\subsection{More Results}

Extensive comparative results of Fig.~\ref{visual_white}, Fig.~\ref{visual_black} and Fig.~\ref{visual_adv} validate the superiority of SEAR in generating imperceptible and transferable adversarial perturbations to deceive the forensic models under three challenging settings. In addition, we have visualized some samples of anti-forensic images predicted by our method in Fig.~\ref{performance}. We can observe that our approach generates the realistic anti-forensics images that fool the forensic models. As shown, Mantra-Net and SPAN can predict the precise masks of the tampered areas. Our SEAR synthesizes the anti-forensic images, which look similar to the original forged images with negligible visual loss. We can see significant changes when they detect the anti-forensic images. The predicted masks of the anti-forensic images mis-localize the tampered regions and even detect no tampered regions. The presented results indicate that our concealer hides the artifacts successfully.

\section{Conclusion}

In this paper, we introduce the novel task of pixel-level anti-forensics and propose a unified framework, SEAR, to combat forgery localization algorithms and evade detection at the pixel level. SEAR employs a concealer to generate perturbation maps and add them to original forged images, producing anti-forensic images that obscure tampering traces and deceive forgery localization models. By incorporating self-supervised training, the concealer learns to constrain disturbances to tampered regions and remove high-frequency traces, resulting in visually imperceptible effects on other regions. Additionally, adversarial learning integrates a well-trained forgery localization network into the system, and after a combination of self-supervised and adversarial training, the concealer can slightly modify original forged images and successfully deceive forgery localization models. We provide detailed experimental results to validate SEAR.
This work paves the way for further exploration of anti-forensics. Interestingly, adversarial learning benefits both anti-forensic algorithms and forensic models and could inspire future work on forgery localization. Our future research focuses on improving the robustness of forensic algorithms based on anti-forensic methods.

\bibliographystyle{IEEEtran}
\bibliography{reference}

  \begin{IEEEbiography}
	[{\includegraphics[width=1in,height=1.25in,clip,keepaspectratio]{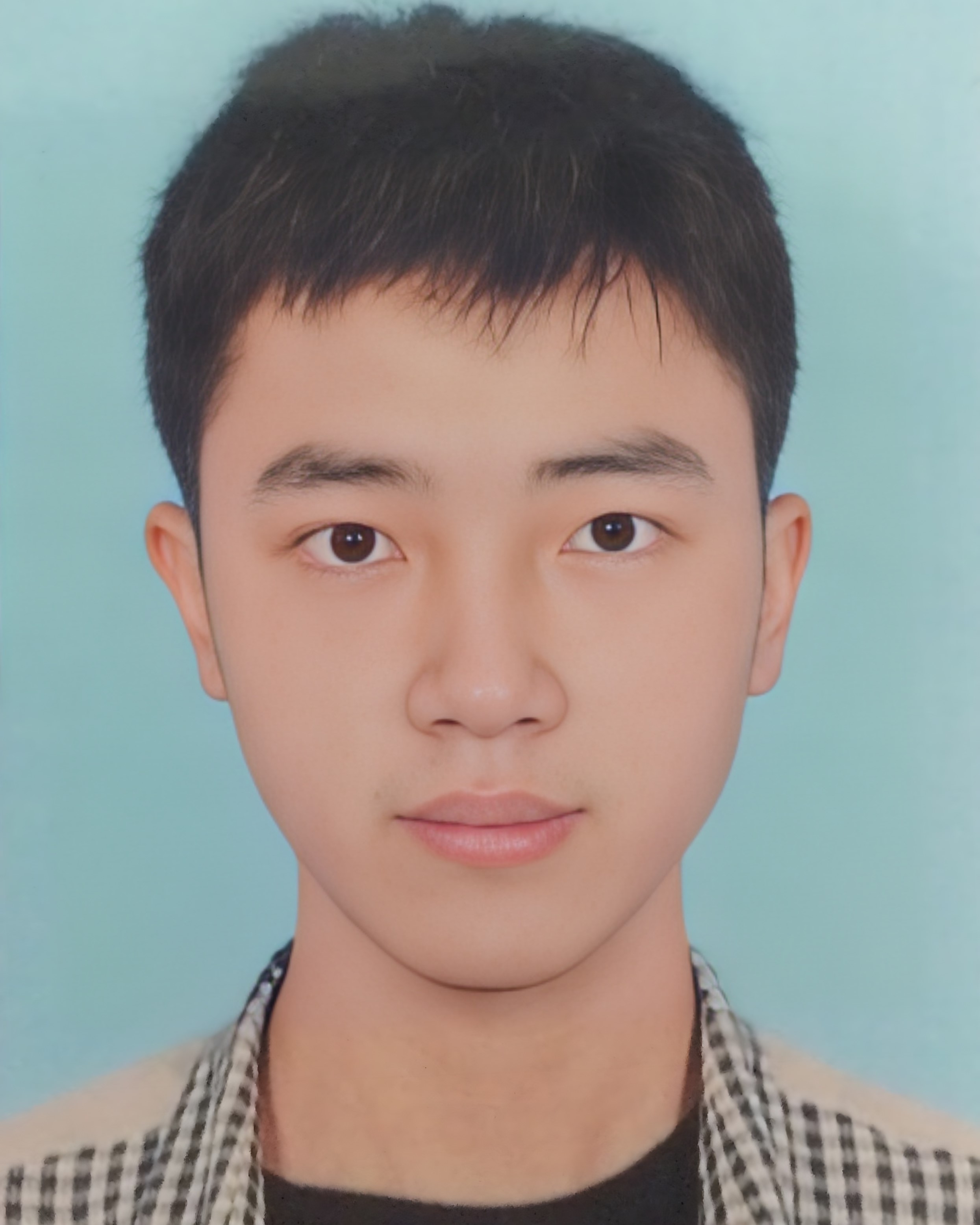}}]{Long
	  Zhuo} received the B.S. degree in College of Computer Science and
	Software Engineering from Shenzhen University, Shenzhen, China in 
	2019.   
	  
	He was a research assistant at the Shenzhen Key Laboratory of Media Security. He is currently a researcher at Shanghai AI Laboratory. His current research interests
	include multimedia forensics, image generation, 3D generation, and video generation.
  \end{IEEEbiography}

  \begin{IEEEbiography}
	[{\includegraphics[width=1in,height=1.25in,clip,keepaspectratio]{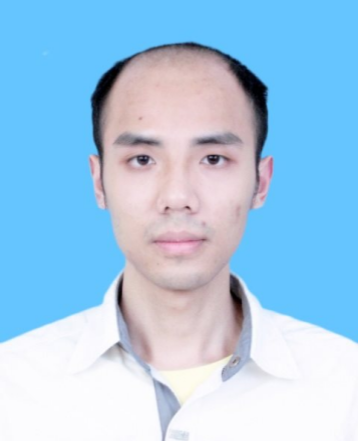}}]{Shenghai Luo} received the B.S. degree in computer science and technology from Southwest University of Science and Technology, China, in 2020.  And he received the master’s degree majoring in computer technology from Shenzhen University, China, in 2023.   His current research interests include multimedia forensics and model compression.

  \end{IEEEbiography}

  \begin{IEEEbiography}[{\includegraphics[width=1in,height=1.25in,clip,keepaspectratio]{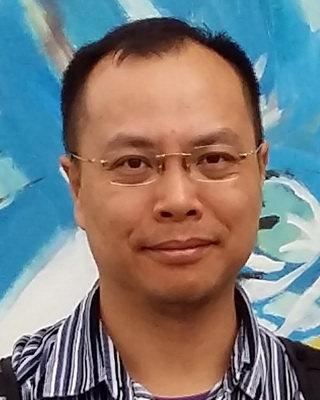}}]{Shunquan Tan (M'10--SM'17)}
	received the B.S. degree in computational mathematics and applied
	software and the Ph.D. degree in computer software and theory from
	Sun Yat-sen University, Guangzhou, China, in 2002 and 2007,
	respectively.
  
	He was a Visiting Scholar with New Jersey Institute of Technology,
	Newark, NJ, USA, from 2005 to 2006. He is currently an Associate
	Professor with College of Computer Science and Software Engineering,
	Shenzhen University, China, which he joined in 2007. He is the Vice
	Director with the Shenzhen Key Laboratory of Media Security. His
	current research interests include multimedia security, multimedia
	forensics, and machine learning.
  \end{IEEEbiography}

  \begin{IEEEbiography}
	[{\includegraphics[width=1in,height=1.25in,clip,keepaspectratio]{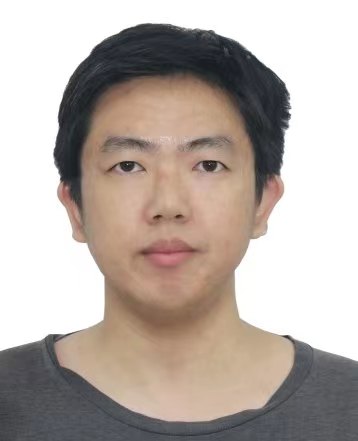}}]{Han Chen} received the B.S. degree in electronic information engineering from Shenzhen University, China, in 2020, where he is currently pursuing the master's degree majoring in information and communication engineering. His current research interests include multimedia forensics and deep learning.

  \end{IEEEbiography}

  \begin{IEEEbiography}[{\includegraphics[width=1in,height=1.25in,clip,keepaspectratio]{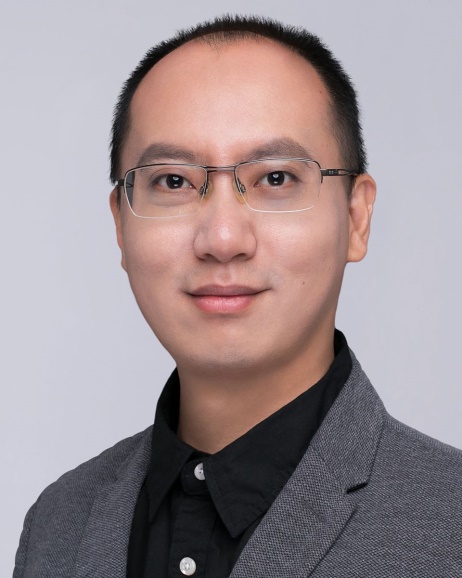}}]{Bin Li (S'07-M'09-SM'17)}
	received the B.E. degree in communication engineering and the Ph.D.
	degree in communication and information system from Sun Yat-sen
	University, Guangzhou, China, in 2004 and 2009, respectively.
  
	He was a Visiting Scholar with the New Jersey Institute of
	Technology, Newark, NJ, USA, from 2007 to 2008. He is currently a
	Professor with Shenzhen University, Shenzhen, China, where he joined
	in 2009. He is also the Director with the Guangdong Key Lab of
	Intelligent Information Processing and the Director with the
	Shenzhen Key Laboratory of Media Security. He is an Associate Editor
	of the IEEE Transactions on Information Forensics and Security. His
	current research interests include multimedia forensics, image
	processing, and deep machine learning.
  \end{IEEEbiography}

  \begin{IEEEbiography}[{\includegraphics[width=1in,height=1.25in,clip,keepaspectratio]{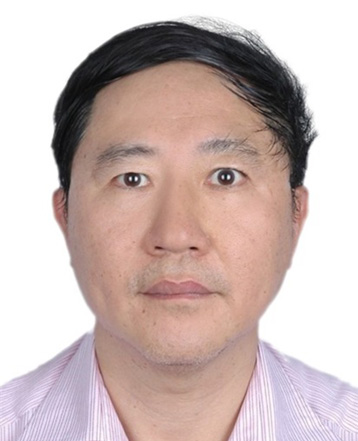}}]{Jiwu Huang (M'98--SM'00--F'16) }
	received the B.S. degree from Xidian University, Xi'an, China, in
	1982, the M.S. degree from Tsinghua University, Beijing, China, in
	1987, and the Ph.D. degree from the Institute of Automation, Chinese
	Academy of Science, Beijing, in 1998. He is currently a Professor
	with the College of Electronics and Information Engineering,
	Shenzhen University, Shenzhen, China. Before joining Shenzhen
	University, he has been with the School of Information Science and
	Technology, Sun Yat-sen University, Guangzhou, China, since
	2000. His current research interests include multimedia forensics
	and security. He is an Associate Editor of the IEEE Transactions on
	Information Forensics and Security.
  \end{IEEEbiography}

\end{document}